\pdfoutput=1
\documentclass[nohyperref]{article}

\usepackage{microtype}
\usepackage{graphicx}
\usepackage{subfigure}
\usepackage{booktabs} 

\usepackage{hyperref}

\usepackage{multibib}

\usepackage[accepted]{icml2022}

\usepackage{amsmath}
\usepackage{amssymb}
\usepackage{mathtools}
\usepackage{amsthm}

\usepackage[capitalize,noabbrev]{cleveref}

\theoremstyle{plain}

\theoremstyle{definition}

\theoremstyle{remark}

\usepackage[textsize=tiny]{todonotes}

\newcites{Appendix}{Reference}

\icmltitlerunning{On the Impact of Knowledge Distillation for Model Interpretability}

\begin{document}

\twocolumn[
\icmltitle{On the Impact of Knowledge Distillation for Model Interpretability}




\begin{icmlauthorlist}
\icmlauthor{Hyeongrok Han}{snu}
\icmlauthor{Siwon Kim}{snu}
\icmlauthor{Hyun-Soo Choi}{snust,ziovision}
\icmlauthor{Sungroh Yoon}{snu,inter}
\end{icmlauthorlist}

\icmlaffiliation{snu}{Department of Electrical and Computer Engineering, Seoul National University, Seoul, Republic of Korea}
\icmlaffiliation{snust}{Department of Computer Science and Engineering, Seoul National University of Science and Technology, Seoul, Republic of Korea}
\icmlaffiliation{ziovision}{ZIOVISION Inc., Chuncheon, Republic of Korea}
\icmlaffiliation{inter}{Interdisciplinary Program in Artificial Intelligence, Seoul National University, Seoul, Republic of Korea}

\icmlcorrespondingauthor{Hyun-Soo Choi}{choi.hyunsoo@seoultech.ac.kr}
\icmlcorrespondingauthor{Sungroh Yoon}{sryoon@snu.ac.kr}

\icmlkeywords{Machine Learning, ICML}

\vskip 0.3in
]



\printAffiliationsAndNotice{}  

\begin{abstract}
Several recent studies have elucidated why knowledge distillation (KD) improves model performance. 
However, few have researched the other advantages of KD in addition to its improving model performance.
In this study, we have attempted to show that KD enhances the interpretability as well as the accuracy of models.
We measured the number of concept detectors identified in network dissection for a quantitative comparison of model interpretability.
We attributed the improvement in interpretability to the class-similarity information transferred from the teacher to student models.
First, we confirmed the transfer of class-similarity information from the teacher to student model via logit distillation.
Then, we analyzed how class-similarity information affects model interpretability in terms of its presence or absence and degree of similarity information.
We conducted various quantitative and qualitative experiments and examined the results on different datasets, different KD methods, and according to different measures of interpretability.
Our research showed that KD models by large models could be used more reliably in various fields.
\end{abstract}

\section{Introduction}
\label{sec:1}

In knowledge distillation (KD), information is transferred from the teacher to student model, improving the performance of the student model~\cite{hinton2015distilling}.
In general, a student model is a small neural network with a lower learning capacity compared to that of the teacher model.
Many attempts have been made to reduce the size of large models using KD~\cite{CLIP-KD1, CLIP-KD2, GPT-KD1}.
This is because the huge size of large pre-trained models such as CLIP and GPT-3 results in increased resource consumption and inference costs, limiting their usage in downstream applications~\cite{GPT, CLIP}.

Several recent studies have elucidated why KD improves model performance~\cite{yuan2020revisiting, tang2020understanding, zhou2021rethinking}.
However, few studies have researched the other advantages of KD besides its improving model performance.
Through this study, we demonstrated that KD could improve not only the generalization performance of models but also the interpretability, which indicates the reliability of models.

Researchers have attempted to understand the internal decision-making processes of neural networks, which essentially seem to be black boxes~\cite{singla2019explanation, sundararajan2017axiomatic, ribeiro2016should}.
For large models such as CLIP and GPT-3 to be applied to various studies, it is necessary to secure explainability~\cite{needXAI1, needXAI2}.
Many studies consider the interpretability of a model high if the activation is object-centric~\cite{object-centric1, object-centric2, object-centric3}.
In this study, we found that KD promoted the object-centricity of the activation map of student models and thereby enhanced their interpretability.

\begin{figure*}[tbp]
    \centering
    \includegraphics[width=0.7\linewidth]{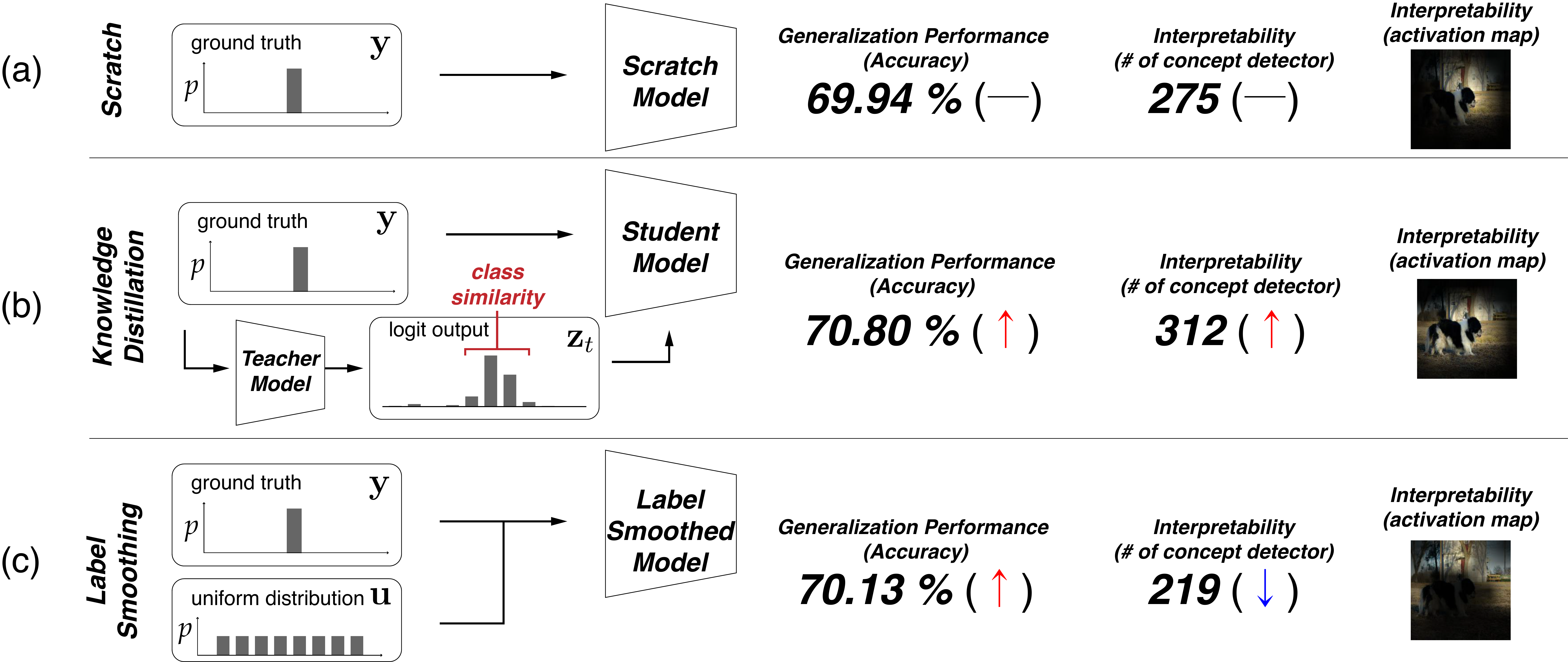}
    
    \caption{Illustration of main argument of the proposed study. The number of concept detectors of different models, namely models trained (a) from scratch ($f_{\mathrm{scratch}}$), (b) using KD ($f_{\mathrm{KD}}$), and (c) using LS ($f_{\mathrm{LS}}$), have been measured for a quantitative comparison of the model interpretability. LS enhances the model performance but reduces the interpretability while KD boosts both. The transfer of class-similarity information from the teacher to student model enhances the model interpretability.}
    \label{fig:main}
\end{figure*}
Figure \ref{fig:main} summarizes the main arguments of this study. 
First, to compare the interpretability of the models, we adopted \textit{the number of concept detectors} introduced in network dissection~\cite{bau2017network} as a measure of interpretability. 
The number of concept detectors represents the degree of the object-centricity of activation maps and is directly proportional to the model interpretability.
According to the defined terms of interpretability, we compared the interpretability of models trained from scratch ($f_{\mathrm{scratch}}$) and trained using KD ($f_{\mathrm{KD}}$), as shown in Figures~\ref{fig:main} (a) and (b).
Comparing the activation maps shown in Figures~\ref{fig:main} (a) and (b), the activation map of $f_{\mathrm{KD}}$ is more object-centric than that of $f_{\mathrm{scratch}}$.

We attributed this improvement in interpretability to the class-similarity information transferred from the teacher to student models.
The distribution of a teacher model had a high similarity between the semantically similar classes. 
For example, when the input image was a Border Collie, the student model was trained to minimize the distance from the distribution of the teacher model $\boldsymbol{z}_t$, which had a high probability of classes belonging to ``dog.''
Thus, whenever ``dog'' samples were used as input, the student model could learn the typical characteristics of a ``dog,'' which supported the object-centricity of the learned representation of the student model.

To demonstrate that class-similarity information enhances the interpretability of student models, we measured the entropy of semantically similar classes to confirm the transfer of class-similarity information from the teacher to student model via logit distillation.
Then, we compared the interpretability between the model trained by label smoothing ($f_{\mathrm{LS}}$), which did not contain (rather negatively affected) class-similarity information, and $f_{\mathrm{KD}}$.
As shown in Figures~\ref{fig:main} (b) and (c), $f_{\mathrm{LS}}$ learns other features than objects, such as the background, which reduces the model interpretability.
Referring to the previous example, for $f_{\mathrm{LS}}$, the probability of an irrelevant class (\textit{e.g.}, valley) increases because the model reduces the distance to a uniform distribution $\boldsymbol{u}$, causing the map of $f_{\mathrm{LS}}$ to become less object-centric.
The results showed that KD enhanced the model performance and interpretability, whereas LS decreased the interpretability.

In addition to analyzing the effect of the presence or absence of class-similarity information on model interpretability, we also analyzed the effect of the \textit{degree} of this information.
\citet{pmlr-v162-chandrasegaran22a} analyzed the effect of a teacher model trained by LS ($f_{\mathrm{LS}}^{\mathrm{teacher}}$) on the performance of the student model.
They showed that the amount of class-similarity information transferred from the teacher to student model could be calibrated by adjusting the temperature of the KD using $f_{\mathrm{LS}}^{\mathrm{teacher}}$.
Accordingly, we analyzed the effect of $f_{\mathrm{LS}}^{\mathrm{teacher}}$ on the interpretability of student models at different temperatures.
Our results showed that 1) the interpretability of the student model improved with the increase in the class-similarity information learned by the students; and 2) adding logit distillation to feature distillation increased the model interpretability, demonstrating the effect of transferring class-similarity information. 

We empirically showed the consistent effect of KD on model interpretability regardless of the KD method and how the interpretability measurement was defined.
In addition to analyzing the effect of vanilla KD, we showed that the interpretability of models trained with various KD methods increased compared to that of $f_{\mathrm{scratch}}$, as discussed in Section~\ref{sec:3.3}.
Then, we generated a synthesized dataset with the ground truth of the heatmap and measured the five-band-scores proposed by~\citet{tjoa2020quantifying} for $f_{\mathrm{scratch}}$ and $f_{\mathrm{KD}}$, as described in Section~\ref{sec:5.1}.
We measured the DiffROAR score~\cite{shah2021input} and loss gradient~\cite{tsipras2018robustness}, as described in Sections~\ref{sec:5.2} and \ref{sec:5.3}, respectively, to assess the improvement in model interpretability according to various measures of interpretability. 

The main contributions of our study are as follows.
\begin{itemize}
  \item[$\bullet$] To the best of our knowledge, this is the first study to show that KD enhances the interpretability as well as the accuracy of models.
  \item[$\bullet$] A comparison of the interpretability of the LS model without similarity information and student models with different amounts of similarity information learned clearly showed that class-similarity information improved the interpretability of the student models.
  \item[$\bullet$] Various quantitative and qualitative experimental results support that KD improves model interpretability across KD methods, notions, and datasets.
\end{itemize}

\section{Related work}
\label{sec:2}
\subsection{Knowledge distillation}
Under KD, the performance of a student model with a low learning capacity is improved by receiving the output distribution from a large pre-trained teacher model with a high learning capacity~\cite{hinton2015distilling}. 
Several studies have proposed various KD methods, including logit distillation~\cite{hinton2015distilling} and feature distillation~\cite{romero2014fitnets, zagoruyko2016paying, yim2017gift, kim2018paraphrasing, xu2020knowledge, tian2019contrastive}, to improve the performance of student models.

Attention transfer (AT) averages the n-channel features in the intermediate layer of the teacher model without any regressor, allowing student models  to learn attention~\cite{zagoruyko2016paying}.
Factor transfer (FT) uses an auto-encoder to provide concise information that student models can easily understand~\cite{kim2018paraphrasing}.
Self-supervision KD (SSKD) uses self-supervision signals to transfer intrinsic semantics and provide information to students~\cite{xu2020knowledge}.
Contrastive representation distillation (CRD) trains students to maximize the lower bound of mutual information between the representations of two models, as done in contrastive learning~\cite{tian2019contrastive}.

Several studies \cite{yuan2020revisiting, tang2020understanding, zhou2021rethinking} have analyzed how KD enhances the generalization performance of student models.
They have confirmed that KD is an adaptive version of LS, which produces a regularization effect on models.
\citet{yuan2020revisiting} analyzed the relationship between KD and LS and proposed teacher-free KD.
However, they did not explain the additional information provided to student models through KD.
In the proposed study, we compared the models trained by LS and KD.
The results revealed that the class-similarity information transferred by the teacher model promoted student models to capture conceptual representations more effectively.

\subsection{Label smoothing}
LS trains a model using a vector that combines a one-hot vector with a uniform distribution as a label~\cite{szegedy2016rethinking}. 
LS employs regularization during training, thereby improving the generalization performance of a model.
\citet{muller2019does} demonstrated that LS renders each example in the training set equidistant from all other classes by visualizing the penultimate layer representations of the image classifiers. 
They demonstrated that $f_{\mathrm{LS}}^{\mathrm{teacher}}$ worsened the performance of the student model because it erased information about the similarities between teacher logits.

On the other hand, \citet{shen2021is} argued that KD and LS were compatible. 
They demonstrated that $f_{\mathrm{LS}}^{\mathrm{teacher}}$ improved the performance of the student model by increasing the distance between the embeddings of semantically similar classes.
\citet{pmlr-v162-chandrasegaran22a} introduced systematic diffusion and analyzed these contradictory findings.
They demonstrated that systematic diffusion curtailed the performance of the student models trained by $f_{\mathrm{LS}}^{\mathrm{teacher}}$, thereby rendering KD at low temperatures effective. 
They also showed that the knowledge distilled from $f_{\mathrm{LS}}^{\mathrm{teacher}}$ resulted in a loss in class-similarity information with the decrease in temperature.

\subsection{Explainable AI}
Researchers have attempted to explain the reasoning processes of deep neural networks (DNNs). 
Post-hoc approaches interpret a trained model by localizing the attended input pixels~\cite{simonyan2014deep, sundararajan2017axiomatic, zeiler2014visualizing, ribeiro2016should} or generating counterfactual explanations~\cite{goyal2019counterfactual, singla2019explanation}. 
In contrast, several researchers~\cite{chen2018looks, alvarez2018towards} have designed a new explainable architecture in which decision-making is inherently interpretable without any post-hoc explanation. 
Most approaches have demonstrated the interpretability of the proposed \textit{methods} via qualitative visualizations or deteriorations in predictions after the elimination of the most important pixels. 

The quantification of \textit{model} interpretability is relatively underexplored. 
\citet{li2020learning} theoretically defined interpretability as local linearity. 
\citet{barcelo2020model} suggested that the computational complexity required to obtain explanations represents interpretability; the lower the complexity, the higher the interpretability. 
However, this cannot be empirically applied because real data distribution is considered, which results in high computational complexity. 
By contrast, network dissection~\cite{bau2017network} provides an intuitive and efficient method for quantifying the interpretability of DNNs. 
Therefore, we adopted network dissection as a measure of interpretability; the details are provided in the next section.

\section{On the impact of KD for model interpretability}
\label{sec:3}

This section investigates the impact of KD on model interpretability.
Section \ref{sec:3.1} describes the process of defining and quantifying model interpretability.
Section \ref{sec:3.2} compares the interpretabilities of $f_{\mathrm{scratch}}$, $f_{\mathrm{KD}}$, and $f_{\mathrm{LS}}$.
Section \ref{sec:3.3} presents the comparison of the model trained using various KD methods to verify that enhancements are not limited to just those done by vanilla KD. 

\subsection{Interpretability quantification via network dissection}
\label{sec:3.1}

As described in the previous section, inspired by network dissection, we measured the interpretability of the models using the number of concept detectors.
First, we shall describe the broadly and densely labeled (Broden) dataset and explain the process of counting the concept detectors.
This dataset comprises the following datasets: ADE20k~\cite{zhou2017scene}, OpenSurfaces~\cite{bell2014intrinsic}, PASCAL-Context~\cite{mottaghi2014role}, PASCAL-Part~\cite{chen2014detect}, and Describable Textures~\cite{cimpoi2014describing}.
The samples in Broden include objects, scenes, parts, textures, materials, and color concepts. 
Annotation masks for visual concepts are permitted in this dataset. 
All the pixels of a sample were annotated based on the corresponding concept.
Therefore, by comparing the activation map for each unit in a neural network with the annotation mask, a unit aligned with human-interpretable concepts could be obtained~\cite{bau2017network}.
\begin{figure*}[tp]
    \centering
    \includegraphics[width=0.8\linewidth]{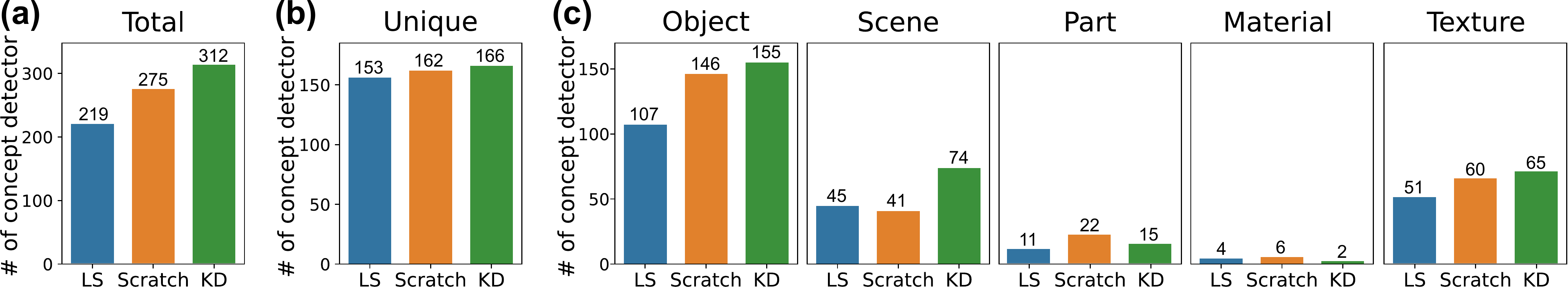}
    \vspace{-2mm}
    \caption{Comparison of interpretabilities among $f_{\mathrm{scratch}}$, $f_{\mathrm{KD}}$, and $f_{\mathrm{LS}}$: (a) Number of concept detectors, (b) unique detectors, and (c) concept detectors per concept; the number of concepts and unique detectors increases for $f_{\mathrm{KD}}$ and decreases for $f_{\mathrm{LS}}$ (particularly, object detector).}
    \label{fig:layer4}
\end{figure*}

The concept detectors were determined as follows.
We measured the interpretability of model $f$ with its frozen weights.
One sample $x$ of Broden was inputted to the model, and the activation map $A_i(x)$ was obtained for the $i$-th convolutional unit, following which the distribution $a_i$ of this map was obtained.
The threshold $T_i$ corresponding to the top 0.5\% of the activation value was calculated from $a_i$ to satisfy $P(a_i \geq T_i) = 0.005$.
Since the activation map $A_i(x)$ had a smaller resolution than the annotation mask $M_c$ did for concept $c$, we interpolated $A_i(x)$ to ensure that the resolutions of $M_c$ and $A_i(x)$ would be identical. 
Subsequently, we performed binary masking on the interpolated $A_i(x)$ such that only regions greater than or equal to $T_i$ would appear.
Finally, we calculated the intersection of union (IoU) scores between the masked activation map and annotation mask $M_c$. 
When the IoU score exceeded a predetermined threshold value (0.05), a unit $i$ was recognized as the concept detector of the corresponding concept $c$.
We have included a pseudocode for obtaining concept detectors in Appendix~\ref{sec:ap_C} to facilitate the understanding of network dissection.

A unique detector is a unit that is aligned with only a single concept.
Network dissection~\cite{bau2017network} reports the number of unique detectors to measure the degree of disentanglement of intermediate representations.
In this study, we measured the number of unique detectors and total number of concept detectors.
We examined the overall concept detection capability of the unit and the degree of disentanglement.

\subsection{Impact of KD on model interpretability}
\label{sec:3.2}
This section investigates the impact of KD on model interpretability, which was defined in the previous section.
The experimental settings for training $f_{\mathrm{scratch}}$, $f_{\mathrm{KD}}$, and $f_{\mathrm{LS}}$ are presented in Appendix~\ref{sec:ap_B.2}.
Figure \ref{fig:layer4} shows the interpretabilities of $f_{\mathrm{scratch}}$, $f_{\mathrm{KD}}$, and $f_{\mathrm{LS}}$.
We measured the number of concept detectors from the last convolutional layer of the model.
A comparison of the interpretabilities for the lower layers of the models is provided in Appendix~\ref{sec:ap_A.2}.
When KD was implemented, the total number of concept detectors, especially object detectors, increased significantly.
Meanwhile, compared with those in $f_{\mathrm{scratch}}$, both the number of concept detectors and unique detectors decreased in $f_{\mathrm{LS}}$; although the number of object detectors decreased significantly, the number of scene detectors increased.
The results are discussed in Section~\ref{sec:4.2}.

Table \ref{tab:number of interpretability} lists the accuracy and interpretability of $f_{\mathrm{scratch}}$, $f_{\mathrm{LS}}$, and $f_{\mathrm{KD}}$.
We verified that KD could improve both the accuracy and interpretability of the models.
To ensure the reliability of our results, we compared the interpretability of various architectures other than ResNet-18.
In addition, we varied the quantile (top-k\%) and IoU threshold, which are essential hyper-parameters for obtaining the concept detector.
The results showed that KD enhanced model interpretability regardless of the architecture and hyper-parameters (these results are presented in Appendices~\ref{sec:ap_A.1} and \ref{sec:ap_A.3}).

\subsection{Verification of various KD methods}
\label{sec:3.3}
In the previous section, we compared the interpretability of $f_{\mathrm{KD}}$ trained using vanilla KD; $f_{\mathrm{scratch}}$; and $f_{\mathrm{LS}}$.
As discussed in Section \ref{sec:2}, various KD methods other than vanilla KD have been proposed.
We verified whether model interpretability improved, even if the teacher model transferred knowledge other than $\boldsymbol{z}_t$ to the student model.
We measured the interpretability of the models trained using various methods, such as AT~\cite{zagoruyko2016paying}, FT~\cite{kim2018paraphrasing}, CRD~\cite{tian2019contrastive}, SSKD~\cite{xu2020knowledge}, and self-KD~\cite{furlanello2018born} (Table~\ref{tab:number of interpretability}).

The accuracy of the model and total number of concept detectors increased regardless of the KD method used.
This implied that when the teacher model transferred knowledge to the student model, its ability to capture the conceptual features of the student model could improve.
Although the number of unique detectors in the AT models decreased compared to that of $f_{\mathrm{scratch}}$, the number of object detectors increased.
In particular, the self-KD model had a significant increase in interpretability; this can be explained as follows.
In general, teacher models have architectures with a higher learning capacity than that of student models. 
Under self-KD, the architectures of both the teacher and student models were the same. 
\citet{cho2019efficacy} argued that if there was a gap between the learning capacities of the teacher and student models, the student model might not effectively understand the content.
Similarly, the interpretability of the student model that received class-similarity information from the ResNet-18 model, which had a similar learning capacity, improved significantly.

\section{Impact of class-similarity information on model interpretability}
\label{sec:4}

\begin{table}[bp]
    \centering
    \vspace{-5mm}
    \caption{Comparison of interpretabilities of various models on layer 4; Acc represents the Top-1 test accuracy on the ImageNet dataset. Each model was trained thrice based on different initial points to avoid variations. The accuracy and interpretability based on the average value of the three models are shown.}
    \vspace{1mm}
    \resizebox{\linewidth}{!}{%
    \begin{tabular}{c| c c c c c c c c c} 
    \toprule
        Model & Object & Scene & Part & Material & Texture & Color & Unique & \textbf{Total} & Acc \\\midrule
        Scratch                   & 146  & 41    & 22 & 6 & 60 & 0 & 162  & 275  & 69.94 \\
        LS           & 107 & 45 & 11    & 4  & 51 & 0 & 153  & 219  & 70.13 \\
        KD (Vanilla)              & 155 & 74 & 15 & 2 & 65 & 1 & \textbf{166}   & \textbf{312}   & \textbf{70.80} \\\midrule
        AT                  & 163 & 35    & 25 & 10 & 54    & 0    & 158    & 287 & 70.52 \\
        FT                  & 162    & 32 & 33    & 11 & 65 & 1    & 172 & 304 & 71.40 \\
        CRD                 & 156 & 34 & 26    & 9  & 57 & 1 & 156    & 283    & 70.68 \\
        SSKD                & 162    & 63    & 15    & 3     & 66    & 1    & 164    & 310    & 70.09 \\
        Self-KD             & 173 & 46 & 22 & 6  & 69 & 0 & 169 & 316    & 70.63  \\\bottomrule
    \end{tabular}
    }
    \label{tab:number of interpretability}
\end{table} 
In this section, we analyze the impact of transferred class-similarity information from the teacher to student model. 
First, in Section~\ref{sec:4.1}, we discuss that class-similarity information is transferred from the teacher to student model via logit distillation. 
Then, we contrasted the interpretability between $f_{\mathrm{LS}}$, which does not contain class-similarity information, and $f_{\mathrm{KD}}$, in Section~\ref{sec:4.2}.
We visualized the activation maps of $f_{\mathrm{scratch}}$, $f_{\mathrm{KD}}$, and $f_{\mathrm{LS}}$, empirically confirming that the activation map became object-centric as class similarity was transferred to the models.
The student interpretability increased as the class-similarity information learned by the student model increased through class-similarity calibration; this is discussed in Section~\ref{sec:4.3}.

\subsection{Examination of provision of class-similarity information by teacher model}
\label{sec:4.1}
To assess whether $f_{\mathrm{KD}}$ contained more class-similarity information than the other models did, we compared the entropy of the entire class and those within the same category.
ImageNet dataset is a hierarchical dataset that comprises 1,000 classes.
First, we divided these classes into 67 categories according to the coarse classification scheme proposed by~\citet{eshed_novelty_detection}.
Among the 67 categories, we excluded the ``other'' category because we could not state that similar classes had been grouped in this category.
We measured the entropy of classes within the same category, which represents the amount of information contained in the model for that category, from the output distribution of $f_{\mathrm{KD}}$.
A larger entropy implied that the model contained more class-similarity information.
The detailed experimental setting for the entropy measurement is presented in Appendix~\ref{sec:ap_B.3}.

Table \ref{tab:class similarity overall class} lists the results of the entropy measurements.
Entropy (entire) was measured based on the output of all 1,000 classes, and entropy (category) was measured based on the output of the classes in the category to which the true class belongs.
For all the classes, the entropy of $f_{\mathrm{LS}}$ was extremely high because the model was trained with a uniform distribution.
By contrast, $f_{\mathrm{LS}}$ had the lowest entropy within the same category; $f_{\mathrm{KD}}$ had the highest entropy.
The results showed that $f_{\mathrm{KD}}$ contained more class-similarity information than the other models did.
\begin{table}[bp]
\centering
    \vspace{-5mm}
    \caption{Comparison of the entropy measured based on the output of all classes (Entire) and output of classes in the same category to which the correct class belongs (Category) for $f_{\mathrm{scratch}}$, $f_{\mathrm{KD}}$, and $f_{\mathrm{LS}}$. Averaged entropy for 1,000 test samples of ImageNet was measured. Averages were only obtained for samples for which all models were correct.}
    \vspace{1mm}
    \resizebox{0.55\linewidth}{!}{%
    \begin{tabular}{c c  c} 
    \toprule
        Model   & Entire & Category  \\\midrule
        Scratch                 & 0.4851            &   2.8986         \\
        KD  & 0.5412            & \textbf{2.9041}  \\
        LS         & \textbf{3.3537}   &   2.6040          \\\bottomrule
    \end{tabular}
    }
    \label{tab:class similarity overall class}
\end{table}
\subsection{Impact of presence of class-similarity information on model interpretability}
\label{sec:4.2}
We compared KD and LS to analyze how class-similarity information affected the degree of object-centricity of an activation map.
First, we compared the loss functions of KD and LS to mathematically analyze the origin of the difference in class-similarity information.
Equations (\ref{eq:kd_loss}) and (\ref{eq:ls_loss}) represent the loss functions of KD ($\mathcal{L}_{KD}$) and LS ($\mathcal{L}_{LS}$), respectively:
\begin{equation}
    \mathcal{L}_{KD} = (1-\alpha) \cdot \mathcal{L}_{CE}(\sigma(\boldsymbol{z}_{s}), \boldsymbol{y})+ \alpha T^2 \cdot \mathcal{L}_{CE}(\sigma(\boldsymbol{z}_{s}^T), \sigma(\boldsymbol{z}_{t}^T)),
    \label{eq:kd_loss}
\end{equation}
\begin{equation}
    \mathcal{L}_{LS} = (1-\alpha) \cdot \mathcal{L}_{CE}(\sigma(\boldsymbol{z}),\boldsymbol{y}) + \alpha \cdot \mathcal{L}_{CE}(\sigma(\boldsymbol{z}), \boldsymbol{u}),
    \label{eq:ls_loss}
\end{equation}
where $\mathcal{L}_{CE}$ denotes the cross-entropy loss; $\sigma$ denotes the softmax function; $\boldsymbol{z}_s$ and $\boldsymbol{z}_t$ represent the output logits (distributions) of the student and teacher models, respectively;
$\boldsymbol{z}$ is the logits of the target model for LS; $\boldsymbol{y}$ denotes the one-hot encoded ground truth vector; and $\boldsymbol{u}$ denotes the uniform distribution; $\alpha$ and $T$ are the hyper-parameters, where $\alpha$ is the mixture parameter and $T$ is the temperature for adjusting the smoothness of the distributions. 
We used the upper index, $T$, for the distributions smoothed by temperature $T$ ($\boldsymbol{z}_s^T$ and $\boldsymbol{z}_t^T$).
When $T = 1$ and the student model was considered the target model, the two loss functions differed only for the distribution $\boldsymbol{z}_t$ of the teacher model and the uniform distribution $\boldsymbol{u}$.
Unlike $\boldsymbol{u}$, $\boldsymbol{z}_t$ contained information regarding the similarity between classes.
Therefore, the difference in interpretability could be potentially attributed to the transfer of class-similarity information to the target model.

\begin{figure}[tbp]
    \centering
    \includegraphics[width=0.9\linewidth]{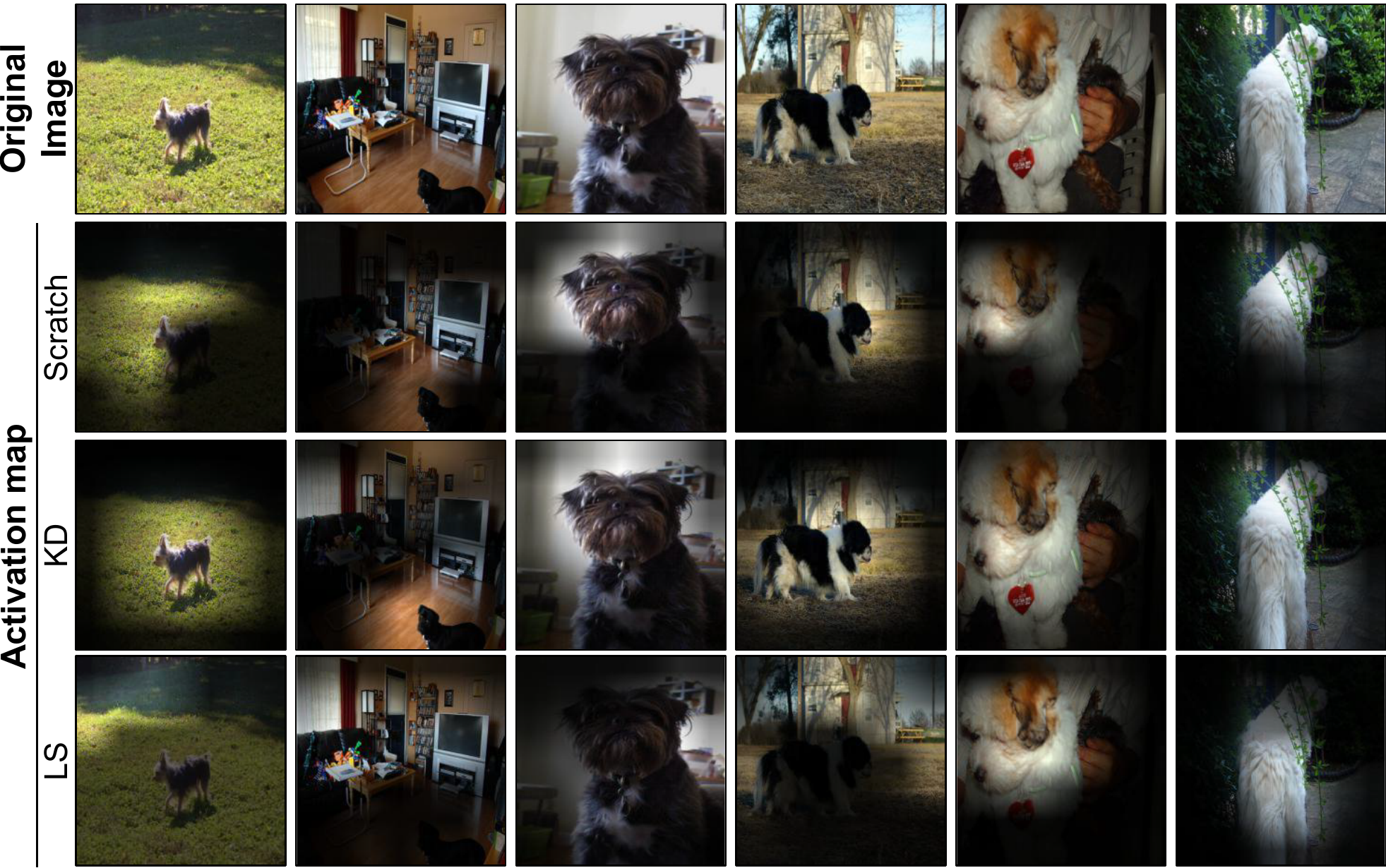}
    \vspace{-3mm}
    \caption{Comparison among activation maps of dog detector under $f_{\mathrm{scratch}}$, $f_{\mathrm{KD}}$, and $f_{\mathrm{LS}}$; the brighter the pixel, the higher the activation value. The dog detector is the concept detector with the highest IoU score on the sample belonging to the ``dog'' concept.
    }
    \label{fig:activation_map}
\end{figure}
We visualized the activation maps generated by the concept detectors; Figure \ref{fig:activation_map} shows the activation maps of $f_{\mathrm{scratch}}$, $f_{\mathrm{KD}}$, and $f_{\mathrm{LS}}$ overlaid on the sample.
We observed that the activation maps of $f_{\mathrm{KD}}$ are more object-centric and activate in the entire object than those of $f_{\mathrm{scratch}}$ and $f_{\mathrm{LS}}$.
We shall explain the improvement in the model interpretability owing to the transferred class-similarity information.
As an input, let us consider the image of a \texttt{Border Collie} belonging to the ``dog'' category.
The logits of classes belonging to the ``dog'' category, such as the \texttt{German Shepherd} and \texttt{Komondor}, are higher than those of other classes that do not belong to the same category in $\boldsymbol{z}_t$.
Even if the image of a \texttt{German Shepherd} was input, the teacher model transferred a similar distribution, which had a high logit of classes in the ``dog'' category, to the student model.
These similar distributions enabled the student models to learn the typical characteristics of dogs from various images of them.
The activation map of the student model could be more object-centric (\textit{e.g.}, torso, ear, tail), as shown in Figure \ref{fig:activation_map}.
Object-centric representation increases the IoU score corresponding to the correct concept.
Meanwhile, $f_{\mathrm{LS}}$ was trained to learn all the other classes (\textit{e.g.}, \texttt{seatbelt} and \texttt{valley}), even when the image of a \texttt{Border Collie} was the input.
The activation map of $f_{\mathrm{LS}}$ was activated in regions independent of the image of the dog. 
This resulted in a lower IoU score for the correct concept.
It can be inferred that KD improved the model interpretability by transferring class-similarity information, whereas LS reduced it.

Next, we shall interpret the experimental results in Table \ref{tab:number of interpretability} and Figure \ref{fig:layer4}.
The total number of concept detectors of $f_{\mathrm{LS}}$ decreased, but the number of scene detectors increased compared with that of $f_{\mathrm{scratch}}$.
Figure \ref{fig:activation_map} shows that the activation map of $f_{\mathrm{LS}}$ is scattered compared with that of the others.
This implied that the activation map had a higher activation value in the scene than the object-centric map did.
The number of object detectors decreased because $f_{\mathrm{LS}}$ captured more locally in the object region, reducing the IoU score compared with that of the other models. 
For $f_{\mathrm{KD}}$, the activation map captured a wider area of the object, including even a part of the background, increasing the number of scene detectors.
The improved interpretability of $f_{\mathrm{KD}}$, decreased interpretability of $f_{\mathrm{LS}}$, and increased number of scene detectors of $f_{\mathrm{LS}}$ can be explained by comparing the visualizations of the activation map.

\subsection{Impact of amount of class-similarity information on model interpretability}
\label{sec:4.3}
\begin{figure}[tp]
    \centering
    \includegraphics[width=0.9\linewidth]{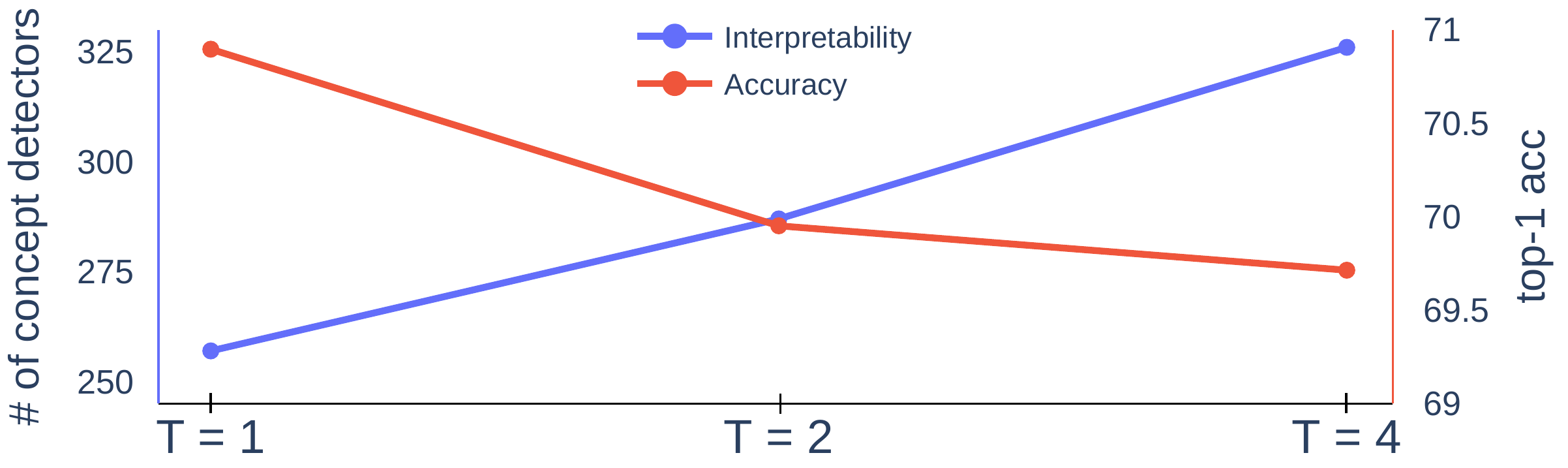}
    \vspace{-3mm}
    \caption{Comparison of interpretability and accuracy of student model trained from $f_{\mathrm{LS}}^{\mathrm{teacher}}$; $f_{\mathrm{LS}}^{\mathrm{teacher}}$ was trained using LS with $\alpha = 0.1$. The accuracy (red line) of students declines as $T$ increases. On the other hand, the interpretability (blue line) of the student models increases as $T$ increases. This result shows that the interpretability of the model improves as it learns class-similarity information better.}
    \label{fig:LS_trained}
\end{figure}
In this section, we shall discuss how the interpretability of the student model improved as the class-similarity information that the student models learned increased from calibrating information.
First, we investigated the effect of $f_{\mathrm{LS}}^{\mathrm{teacher}}$ on the interpretability of student models with varying $T$.
\citet{pmlr-v162-chandrasegaran22a} showed that through KD from $f_{\mathrm{LS}}^{\mathrm{teacher}}$ with a higher $T$, the distances between embeddings belonging to similar and dissimilar classes had become relatively reduced and increased, respectively.
This implied that the higher the $T$, the greater the amount of class-similarity information the student model learned from $f_{\mathrm{LS}}^{\mathrm{teacher}}$, which allowed us to calibrate the class-similarity information that the student model learned. 

Figure \ref{fig:LS_trained} shows the interpretability and accuracy of the student models trained by $f_{\mathrm{LS}}^{\mathrm{teacher}}$. 
The accuracy of these models gradually decreased as $T$ increased, aligning with the analysis reported by \citet{pmlr-v162-chandrasegaran22a}.
On the other hand, the interpretability of these models gradually increased as $T$ increased. 
The detailed settings of KD using $f_{\mathrm{LS}}^{\mathrm{teacher}}$ are presented in Appendix~\ref{sec:ap_B.5}.

\begin{figure}[tbp]
    \centering
    \includegraphics[width=0.9\linewidth]{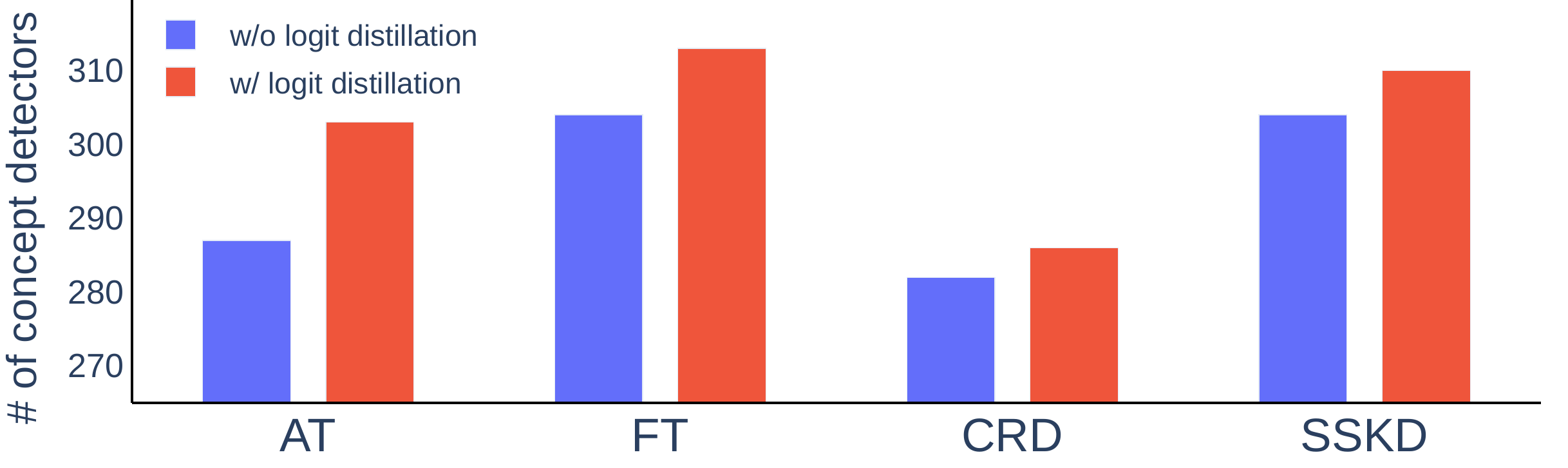}
    \vspace{-3mm}
    \caption{Comparison of interpretability between student models with (red bar) and without (blue bar) logit distillation added to feature distillation. For logit distillation, $T$ and $\alpha$ are 4 and 0.5, respectively. We confirmed that adding logit distillation to feature distillation improved the interpretability for all KD methods.}
    \label{fig:feature distillation}
\end{figure}

Second, we examined the impact of adding logit distillation to feature distillation on model interpretability.
As shown in Table~\ref{tab:number of interpretability}, the interpretability of student models trained using various KD methods other than logit distillation (vanilla KD) also improved.
This aligned with our insight that the information transferred to the student models improved their interpretability.
Intuitively, $\boldsymbol{z}_t$ contained more class-similarity information than features.
We compared the interpretability of student models that learned more class-similarity information by adding logit distillation to feature distillation with that of student models without this addition.
Figure~\ref{fig:feature distillation} shows that interpretability improved when feature distillation was combined with logit distillation.
These results supported our argument that model interpretability improved when models learned more class-similarity information.

\section{Improvements in different measures of model interpretability through KD}
\label{sec:5}
The improvement in measures of interpretability other than the number of concept detectors, namely five-band-scores, DiffROAR scores, and loss gradient, shall be discussed.
\citet{tjoa2020quantifying} proposed five-band-scores to measure model interpretability by using a synthesized dataset with the ground truth of the heatmap.
\citet{shah2021input} proposed the DiffROAR score, a metric used to probe whether an instance-specific explanation of a model highlighted its discriminative features.
\citet{tsipras2018robustness} claimed that the degree of alignment between pixels more relevant to human perception and gradients indicated their degree of interpretability.
We also show that KD improves model interpretability, not only in specific domains (vision), but also across diverse domains (NLP).
In addition to the number of concept detectors, we demonstrated that consistent results could be obtained for various notions, datasets, and domains.

\subsection{Five-band-scores}
\label{sec:5.1}
We measured the interpretability of the KD models using a dataset other than Broden.
For an objective and quantitative evaluation, we used a synthesized dataset with the ground truth for the heatmap proposed by ~\citet{tjoa2020quantifying}.
The examples for each class and the ground truth generated are shown in Appendix~\ref{sec:ap_D}. 
There are three regions on the ground truth of the synthesized dataset: class 0) a background without any classification information (shown as a white region); class 1) localization information, which describes the location of an object (light pink region); and class 2), the distinguishing feature, which is crucial for distinguishing classes (dark pink area).
We trained the model from scratch and KD using this synthesized dataset; details regarding the training are provided in Appendix~\ref{sec:ap_B.6}.
\begin{table}[tbp]
\centering
\vspace{-2mm}
\caption{Comparison of five-band-scores of $f_{\mathrm{scratch}}$ and $f_{\mathrm{KD}}$ on the synthesized dataset. Higher values of pixel accuracy, recall, and precision and lower values of FPR indicate that the model has a higher interpretability; $f_{\mathrm{KD}}$ has a higher interpretability than $f_{\mathrm{scratch}}$ does for all the metrics.}
\vspace{1mm}
\resizebox{1.0\linewidth}{!}{%
    \begin{tabular}{c| c  c  c  c }  
    \toprule
        Model      & Pixel\_acc($\uparrow$) & Recall($\uparrow$) & Precision($\uparrow$) & FPR($\downarrow$)  \\\midrule
        Scratch    & 0.8803  & 0.5014 & 0.3011 & 0.1911 \\
        KD         & \textbf{0.8974} & \textbf{0.5770}  & \textbf{0.3545} & \textbf{0.1871} \\\bottomrule
    \end{tabular}
    }
    \label{tab:five band score}
\end{table} 
\begin{figure}[tbp]
    \centering
    \includegraphics[width=1.0\linewidth]{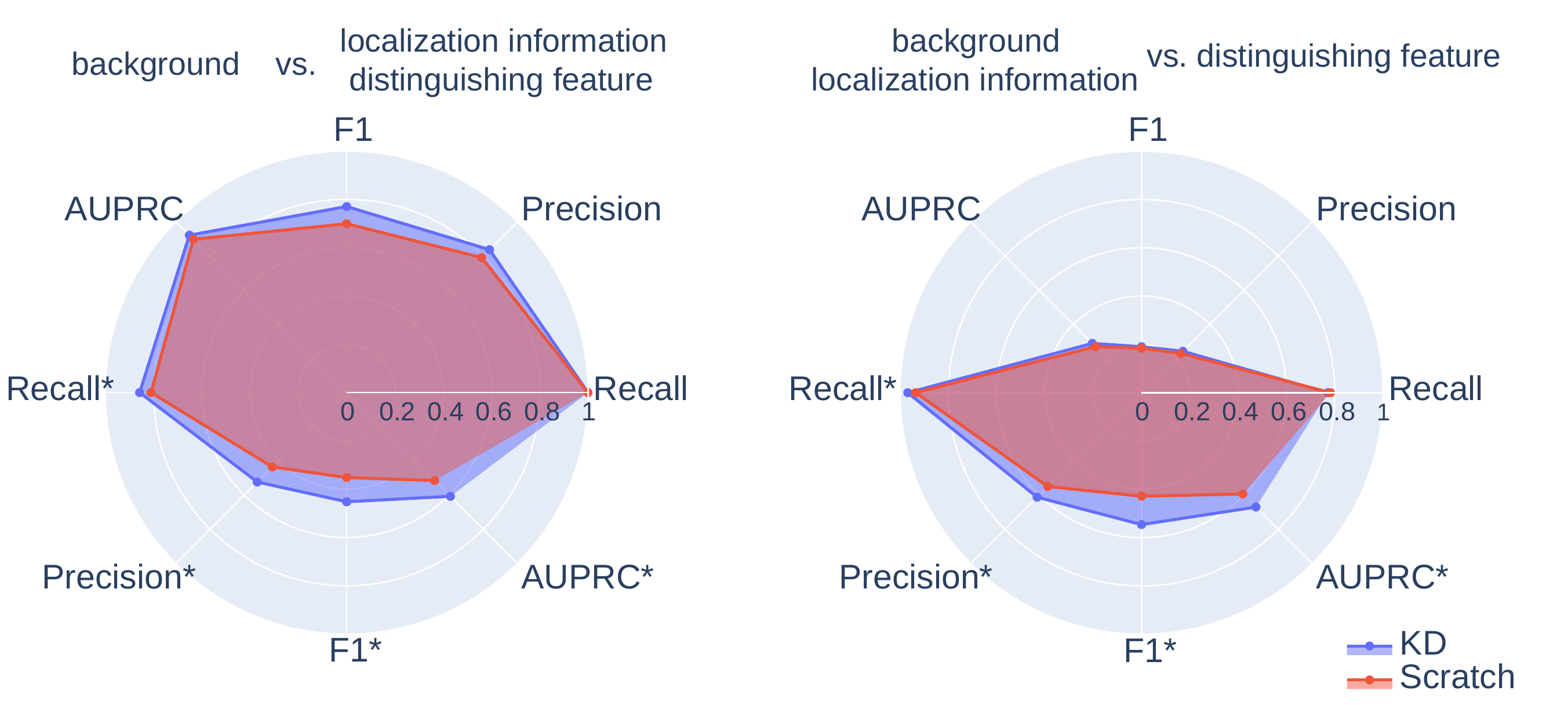}
    \vspace{-2mm}
    \caption{Results based on synthesized dataset. Left: class 0 vs. classes 1,2; Right: classes 0,1 vs. class 2; and * indicates the maximum value for the metric. In both scenarios, $f_{\mathrm{KD}}$ has a higher interpretability than $f_{\mathrm{scratch}}$ does for the averaged and maximum scores of AUROC, Precision, Recall and F1.}
    \label{fig:precision recall}
\end{figure}
\vspace{-2mm}

Since the distinguishing features could be regarded as the correct answer for interpretation, we could measure the interpretability by performing a pixel-by-pixel comparison of the saliency map and ground truth.
However, obtaining the general recall and precision values was difficult because the ground truth was a ternary class rather than a binary class. 
Since the five-band-score reflects pixel accuracy, recall, precision, and the false positive rate (FPR) based on ternary classification, we first measured the five-band-scores for the two models; Table \ref{tab:five band score} shows the results.
In addition, we converted the ternary classification task to a binary classification task. Subsequently, we measured the performance by considering the localization information and distinguishing features as one class and the background and localization information as one class; the results are shown in Figure \ref{fig:precision recall}. 
The left radar plot shows the experimental results for class 0 vs. classes 1 and 2, while the one on the right shows the results for classes 0 and 1 vs. class 2.
We measured the average precision, recall, AUPRC, and F1 scores for $f_{\mathrm{scratch}}$ and $f_{\mathrm{KD}}$.
When reporting the binary classification performance, we excluded samples belonging to a class with no objects (last column in Figure 11) from the evaluation.
Both Table \ref{tab:five band score} and Figure \ref{fig:precision recall} show that the interpretability of $f_{\mathrm{KD}}$ is higher than that of $f_{\mathrm{scratch}}$.

\begin{figure}[tbp]
    \centering
    \includegraphics[width=0.9\linewidth]{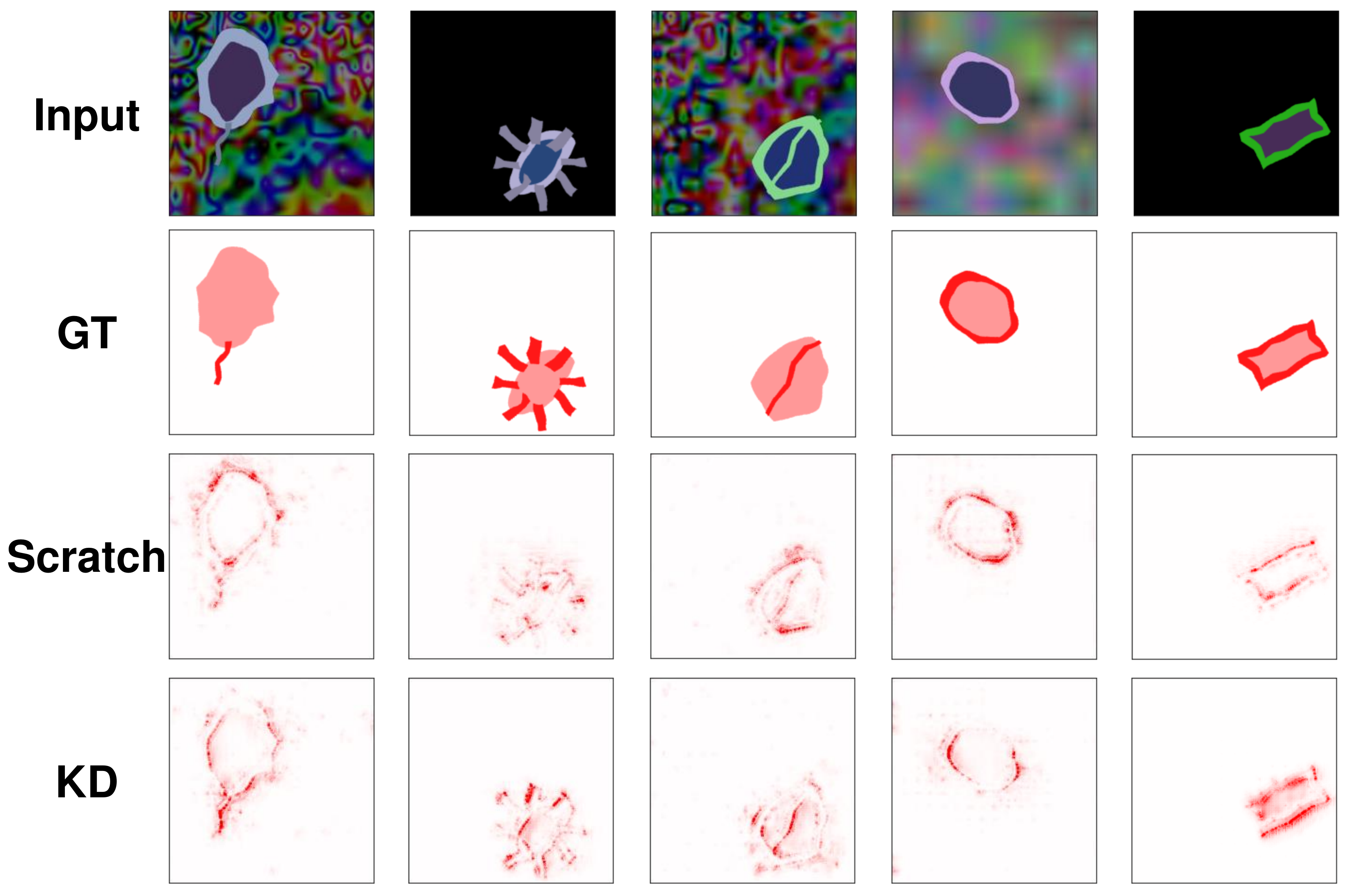}
    \vspace{-3mm}
    \caption{Qualitative results based on synthesized dataset; the first and second rows represent the input image and corresponding ground truth for each sample, respectively, while the third and last rows show the saliency maps of $f_{\mathrm{scratch}}$ and $f_{\mathrm{KD}}$, respectively.}
    \label{fig:synthesized dataset}
\end{figure}

For the qualitative evaluation, we visualized the heatmaps of $f_{\mathrm{scratch}}$ and $f_{\mathrm{KD}}$, as shown in Figure \ref{fig:synthesized dataset}.
We discovered that the saliency map of $f_{\mathrm{scratch}}$ was activated more significantly in the region of the localization information than it had in the region of the distinguishing feature. 
In contrast, the saliency map of $f_{\mathrm{KD}}$ was activated more significantly for the distinguishing features.
The hierarchical structure of the synthesized samples resulted in the synthesized dataset containing similarity information between the classes.
The pre-trained teacher model provided class-similarity information to the student models, which improved their ability to capture the distinguishing features.
Using a synthetic dataset, we showed that KD could improve the model interpretability within various datasets.
\begin{table}[tp]
\vspace{-1mm}
\caption{DiffROAR scores on various datasets (the higher, the better). Nine DiffROAR scores were measured for each dataset by varying the masking fractions (from 0.1 to 0.9 in increments of 0.1). Measurement was repeated with ten different initialization settings. Each value in the table is the average DiffROAR score of these 90 measurements.}
\centering
\vspace{1mm}
\resizebox{0.79\linewidth}{!}{%
    \begin{tabular}{c| c  c  c }  
    \toprule
        Dataset    &  CIFAR-100 & CIFAR-10 & MNIST  \\\midrule
        Scratch    &  3.9747 & 3.3873 & 18.3628     \\
        KD         &  \textbf{4.2001}  & \textbf{4.3850} & \textbf{22.3209} \\
        \bottomrule
    \end{tabular}
    }
    \label{tab:DiffROAR}
\end{table}

\subsection{DiffROAR}
\label{sec:5.2}
DiffROAR is the difference in the predictive power of the datasets, with the top and bottom k\% of the pixels removed by ordering the feature attribution of the model.
A higher DiffROAR score implies that the attributes of the model are well aligned with the task-relevant features.
We measure the DiffROAR scores of $f_{\mathrm{scratch}}$ and $f_{\mathrm{KD}}$ for the CIFAR-100, CIFAR-10, and MNIST test sets.
The results are in Table \ref{tab:DiffROAR}.
For all the datasets, $f_{\mathrm{KD}}$ had a higher DiffROAR score than $f_{\mathrm{scratch}}$ did.

\subsection{Loss gradients}
\label{sec:5.3}
\begin{figure}[tp]
    \centering
    \includegraphics[width=0.9\linewidth]{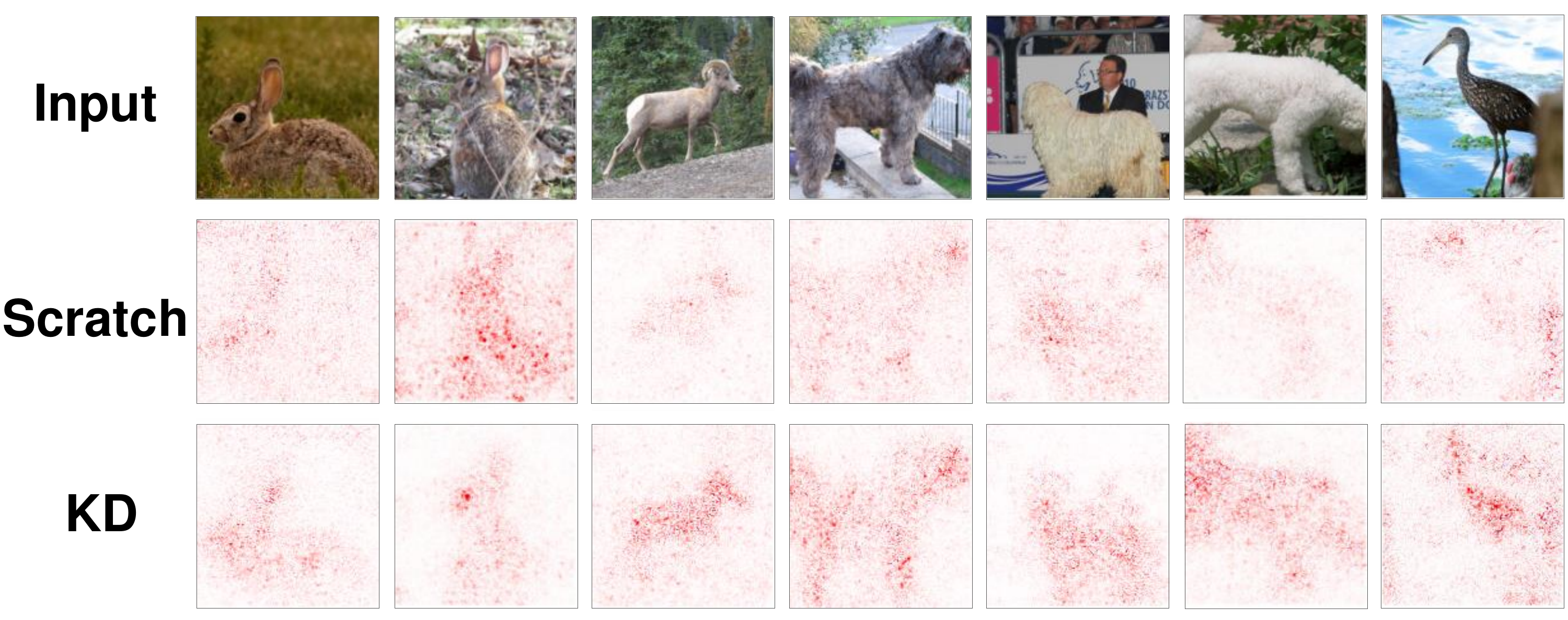}
    \vspace{-3mm}
    \caption{Visualization of loss gradients of $f_{\mathrm{scratch}}$ and $f_{\mathrm{KD}}$ on the test set of ImageNet. Gradients of $f_{\mathrm{KD}}$ are more aligned with the semantically meaningful regions than those of $f_{\mathrm{scratch}}$ are, showing that $f_{\mathrm{KD}}$ learned more human-perceptually relevant features than $f_{\mathrm{scratch}}$.}
    \label{fig:Loss gradient_ImageNet}
\end{figure}

Figure \ref{fig:Loss gradient_ImageNet} shows the loss gradients for the input pixels of $f_{\mathrm{scratch}}$ and $f_{\mathrm{KD}}$ for the ImageNet dataset; the gradients of $f_{\mathrm{KD}}$ are more aligned with the salient characteristic (i.e., the edge of the object) than those of $f_{\mathrm{scratch}}$, showing that $f_{\mathrm{KD}}$ learned more human-perceptionally relevant features than $f_{\mathrm{scratch}}$ did.
The results of the DiffROAR and loss gradient experiments revealed that KD enhanced model interpretability across various interpretability notions, in addition to the number of concept detectors.

\subsection{NLP distillation}
\label{sec:5.4}
In this section, we conduct an experiment using BERT~\cite{devlin2018bert} model for a text classification task, we demonstrated that KD enhances model interpretability in NLP tasks.
To measure model interpretability in line with the convention in NLP, we used the Standard Sentiment Treebank (SST) dataset~\cite{hase2021out, yin2021sensitivity, bastings2021will}.
Below is the protocol we followed to evaluate the interpretability of an NLP model using the SST dataset.
\vspace{-2mm}
\begin{itemize}
    \item[$\bullet$] Data description
    \begin{enumerate}
        \item[$\circ$] Input: sentences (movie review)
        \item[$\circ$] Class: 4-class (`very negative', `negative', `positive', and `very positive')
    \end{enumerate}
    \item[$\bullet$] Model description
    \begin{enumerate}
        \item[$\circ$] Teacher model: 12-layer BERT (acc: 0.623)
        \item[$\circ$] Scratch model ($f_{\mathrm{scratch}}$): 3-layer BERT (acc: 0.484)
        \item[$\circ$] Student model ($f_{\mathrm{KD}}$): 3-layer BERT (\textbf{acc: 0.516})
    \end{enumerate}
    
\end{itemize}

It is noteworthy that the similarity between `very negative' and `negative' as well as between `very positive' and `positive' can be considered class-similarity information.
The SST dataset provides a label for each word as either positive or negative, which serves as the ground truth for saliency (attribution), similar to the synthesized dataset in section~\ref{sec:5.1}.
Therefore, we evaluated how well the saliency (attribution) obtained from $f_{\mathrm{scratch}}$ and $f_{\mathrm{KD}}$ aligns with the ground truth of attribution.
We computed Integrated Gradients (IG) attribution scores from the validation and test samples. 

To quantitatively measure model interpretability, we followed the process outlined below for samples where the model correctly predicts the answer: 1) For samples labeled as `very positive/positive' in sentiment, we measured whether the words labeled as `positive' have positive attribution scores and the words labeled as `negative' have negative attribution scores. 2) For samples labeled as `very negative/negative' in sentiment, we measured whether the words labeled as `negative' have positive attribution scores and the words labeled as `positive' have negative attribution scores.
Table~\ref{tab:NLP distillation} shows the \textit{interpretability} of $f_{\mathrm{scratch}}$ and $f_{\mathrm{KD}}$.
We show the average value of the models trained three times with different initial point.
Through the above experiment, we demonstrated that KD can enhance model interpretability not only in image classification but also in the NLP domain.
The detailed experimental settings of NLP distillation and the results for various model are presented in the Appendix.
\vspace{-3mm}
\begin{table}[h]
\vspace{-1mm}
\caption{Comparison of model interpretability of $f_{\mathrm{scratch}}$ and $f_{\mathrm{KD}}$ on the SST dataset (the higher the better); $f_{\mathrm{KD}}$ has a higher interpretability than $f_{\mathrm{scratch}}$ does for NLP tasks.}
\centering
\vspace{1mm}
\resizebox{0.8\linewidth}{!}{%
    \begin{tabular}{c| c  c  c }  
    \toprule
        Model                   &  Accuracy        & AUROC          & AUPRC  \\\midrule
        $f_{\textrm{scratch}}$  &  0.677           & 0.689          & 0.810     \\
        $f_{\textrm{KD}}$       &  \textbf{0.722}  & \textbf{0.720} & \textbf{0.831} \\
        \bottomrule
    \end{tabular}
    }
    \label{tab:NLP distillation}
\end{table}
\vspace{-5mm}

\section{Conclusions}
\label{sec:7}
In this study, we demonstrated that KD could improve both the interpretability and accuracy of models.
We measured the number of concept detectors of $f_{\mathrm{scratch}}$, $f_{\mathrm{KD}}$, and $f_{\mathrm{LS}}$ to quantitatively compare their interpretabilities. 
The results showed that the number of concept detectors had been significantly increased in $f_{\mathrm{KD}}$ and that the activation of $f_{\mathrm{KD}}$ was more object-centric than those of $f_{\mathrm{scratch}}$ and $f_{\mathrm{LS}}$ were.
We attributed this improvement in interpretability to the class-similarity information transferred from the teacher to student model.
Comparing the interpretability of models with and without class-similarity information showed that class-similarity information had improved the interpretability of the student model.
Additionally, it was revealed that interpretability of the student models improved as the class-similarity information they learn increased with the calibration of information.
Then, we measured the interpretability of the models trained using various KD methods and empirically showed their improved interpretability and performance. 
In addition to the Broden dataset, we measured the interpretability of the synthesized dataset using the ground truth label of the heatmap, the DiffROAR scores, and the loss gradients. 
The consistent results for varying measures of interpretability, KD methods, and datasets supported our argument that KD enhanced model interpretability.

\noindent \textbf{Future scope of this paper} With the emergence of foundation models such as CLIP and GPT, AI is increasingly becoming integrated into various aspects of human life. 
To democratize large foundation models at a reasonable cost and with fewer resources, it is essential that KD becomes prevalent in the future. Considering the concerns regarding the reliability of AI in human life, we believe that our discovery of KD improving model interpretability in crucial areas such as vision and NLP is envisioning. 
Furthermore, more interpretable models make debugging easier, so our findings are also important for engineers from a model development perspective. 
In particular, one possible application direction of our work is demonstrated by the fact that even simple techniques, such as Self-KD, can make the model more interpretable, facilitating easier debugging and improvement.
We are envisioning further studies to determine whether KD can yield other useful results (\textit{e.g.}, the robustness of the model) and whether additional metrics can be used to measure model interpretability.

\section*{Acknowledgement}
This work was supported by the National Research Foundation of Korea (NRF) grants funded by the Korea government (Ministry of Science and ICT, MSIT) (2022R1A3B1077720 and 2022R1A5A708390811), Institute of Information \& communications Technology Planning \& Evaluation (IITP) grants funded by the Korea government (MSIT) (2021-0-02068, 2022-0-00959, and 2021-0-01343: AI Graduate School Program, SNU), Basic Science Research Program through NRF funded by the Korea government (Ministry of Education) (2022R1F1A1076454), and the BK21 FOUR program of the Education and Research Program for Future ICT Pioneers, Seoul National University in 2023.

\nocite{langley00}

\bibliography{mybib}
\bibliographystyle{icml2022}

\newpage
\appendix
\onecolumn

This document is a supplement to our study titled \textbf{`On the Impact of Knowledge Distillation for Model Interpretability'}.
We demonstrate that $f_{\mathrm{KD}}$ improves model interpretability compared to $f_{\mathrm{scratch}}$ across various settings, including different hyper-parameters, layers and architectures.
To ensure reproducibility, we provide a detailed description of the experimental settings used in the main paper.
Additionally, we offer pseudocode for obtaining concept detectors, which aids in further understanding network dissection.
Through entropy measurement within two similar classes, we demonstrate that $f_{\mathrm{KD}}$ contains more class-similarity information than $f_{\mathrm{LS}}$.
In addition to ImageNet, we present the visualization of loss gradients for $f_{\mathrm{scratch}}$ and $f_{\mathrm{KD}}$ on the MNIST dataset.

\section{On the impact of KD for model interpretability in various settings}
\label{sec:ap_A}
We conducted a comparative analysis of model interpretability using a vanilla KD method across various settings.
In Section \ref{sec:ap_A.1}, we present the model interpretability results considering different hyper-parameters, namely $\alpha$, $T$, top k\% activation value and IoU threshold.
Section \ref{sec:ap_A.2} provides a comparison of the interpretability between $f_{\mathrm{scratch}}$, $f_{\mathrm{KD}}$ and $f_{\mathrm{LS}}$ specifically focusing on the lower layers.
In Section \ref{sec:ap_A.3}, we evaluate the model interpretability for various architectures other than ResNet-18.
Our experimental results consistently align with the results presented in the main paper.

\subsection{Model interpretability for various hyper-parameters}
\label{sec:ap_A.1}
In accordance with Equation (1) presented in the main paper, the KD loss function ($\mathcal{L}_{KD}$) incorporates hyper-parameters $\alpha$ and temperature $T$. 
We provide the number of concept detectors for each concept, the number of unique detectors, the total number of concept detectors and the top-1 accuracy on the ImageNet for $f_{\mathrm{KD}}$, with variations of $\alpha$, in Table \ref{tab:kd_alpha_0.1} ($\alpha=0.1$), Table \ref{tab:kd_alpha_0.5} ($\alpha=0.5$) and Table \ref{tab:kd_alpha_0.9} ($\alpha=0.9$).
Within each table, we demonstrate the model interpretability by varying the temperature by values of 1, 4, 8, and 16.
Consistent with the main paper, the presented results are averages from three separate model training runs initiated from different starting point.

When $\alpha$ was set to 0.9, the test accuracy of the other models, except for the case where $T$ was set to one, was lower than that of $f_{\mathrm{scratch}}$.
Teacher output probability of incorrect class compared to the correct class increases as the $\alpha$ value increases.
Training the student model to minimize the distance of the teacher distribution, which was greatly smoothed by a high temperature, resulted in more instances where the student learned information from a class that was not the correct answer.
Consequently, we can explain that when both $\alpha$ and $T$ values were high, the test accuracy of $f_{\mathrm{KD}}$ could be lower than that of $f_{\mathrm{scratch}}$.\\
\begin{table}[t]
\centering
\caption{Interpretability and accuracy of $f_{\mathrm{KD}}$ with $\alpha = 0.1$}
\resizebox{0.7\linewidth}{!}{%
\begin{tabular}{cccccccccc}\toprule
Temperature($T$) & Object & Scene & Part & Material & Texture & Color & Unique & Total & Acc   \\\midrule
1           & 154    & 40    & 23   & 7        & 59      & 0     & 166    & 285   & 70.61 \\
4           & 146    & 47    & 21   & 7        & 70      & 0     & 163    & 292   & 70.62 \\
8           & 147    & 46    & 18   & 4        & 70      & 0     & 158    & 285   & 70.64 \\
16          & 156    & 48    & 19   & 7        & 65      & 0     & 164    & 295   & 70.64 \\ \bottomrule
\end{tabular}
}
\label{tab:kd_alpha_0.1}
\vspace{3mm}
\caption{Interpretability of $f_{\mathrm{KD}}$ with $\alpha = 0.5$}
\resizebox{0.7\linewidth}{!}{%
\begin{tabular}{cccccccccc}\toprule
Temperature($T$) & Object & Scene & Part & Material & Texture & Color & Unique & Total & Acc   \\\midrule
1           & 150    & 43    & 19   & 6        & 58      & 0     & 156    & 277   & 71.17 \\
4           & 155    & 74    & 15   & 2        & 65      & 1     & 166    & 311   & 70.80 \\
8           & 161    & 70    & 21   & 4        & 72      & 1     & 168    & 330   & 70.77 \\
16          & 167    & 71    & 20   & 5        & 79      & 0     & 171    & 343   & 70.85 \\ \bottomrule
\end{tabular}
}
\label{tab:kd_alpha_0.5}
\vspace{3mm}
\caption{Interpretability and accuracy of $f_{\mathrm{KD}}$ with $\alpha = 0.9$}
\resizebox{0.7\linewidth}{!}{%
\begin{tabular}{cccccccccc}\toprule
Temperature($T$) & Object & Scene & Part & Material & Texture & Color & Unique & Total & Acc   \\\midrule
1           & 162    & 48    & 24   & 7        & 69      & 0     & 174    & 310   & 70.87 \\
4           & 137    & 78    & 13   & 4        & 66      & 1     & 166    & 298   & 69.59 \\
8           & 149    & 61    & 20   & 6        & 81      & 0     & 170    & 318   & 68.91 \\
16          & 152    & 72    & 18   & 3        & 79      & 1     & 160    & 325   & 69.54 \\ \bottomrule
\end{tabular}
}
\label{tab:kd_alpha_0.9}
\vspace{3mm}
\caption{Interpretability and accuracy of $f_{\mathrm{LS}}$ with various $\alpha$}
\resizebox{0.7\linewidth}{!}{%
\begin{tabular}{cccccccccc}\toprule
$\alpha$ & Object & Scene & Part & Material & Texture & Color & Unique & Total & Acc   \\\midrule
0.1           & 107    & 45    & 11   & 4        & 51      & 0     & 150    & 219   & 70.01 \\
0.5           & 109    & 81    & 4    & 4        & 56      & 0     & 153    & 255   & 68.01 \\ \bottomrule
\end{tabular}
}
\label{tab:lsr}

\end{table}

We verified that the total number of concept detectors in $f_{\mathrm{KD}}$ increased compared to $f_{\mathrm{scratch}}$, regardless of the combination of hyper-parameters.
Additionally, we demonstrated that the number of object detectors in $f_{\mathrm{KD}}$ increased due to the class-similarity information provided by the teacher, enabling the student to learn more object-centric representations.
These results support our claim that KD enhances model interpretability.
Moreover, in most cases, $f_{\mathrm{KD}}$ exhibited a higher number of unique detectors compared to $f_{\mathrm{scratch}}$, indicating better capture of disentangled representations.
If the $\alpha$ value remained the same, we observed an improvement in model interpretability as the temperature increased in the majority of cases.
This improvement can be attributed to the increased transfer of class-similarity information from the teacher to the student model as the temperature increased.

Table \ref{tab:lsr} lists the number of concept detectors per concept, the number of unique detectors, the total number of concept detectors, and test accuracy of the $f_{\mathrm{LS}}$ with different $\alpha$ values.
We trained the models with two $\alpha$ values (0.1 and 0.5). 
Compared to $f_{\mathrm{scratch}}$, both models exhibited a decrease in interpretability.
Specifically, the number of object detectors significantly decreased, while the number of scene detectors increased compared to $f_{\mathrm{scratch}}$.
This can be attributed to $f_{\mathrm{LS}}$ being trained to minimize the distance with the uniform distribution, enabling it to learn all other classes in addition to the target class.
Consequently, the activation map of $f_{\mathrm{LS}}$ showed more active in the scene surrounding the object rather than at the center of the object. 

\begin{figure}[t]
    \centering
    \includegraphics[width=0.7\linewidth]{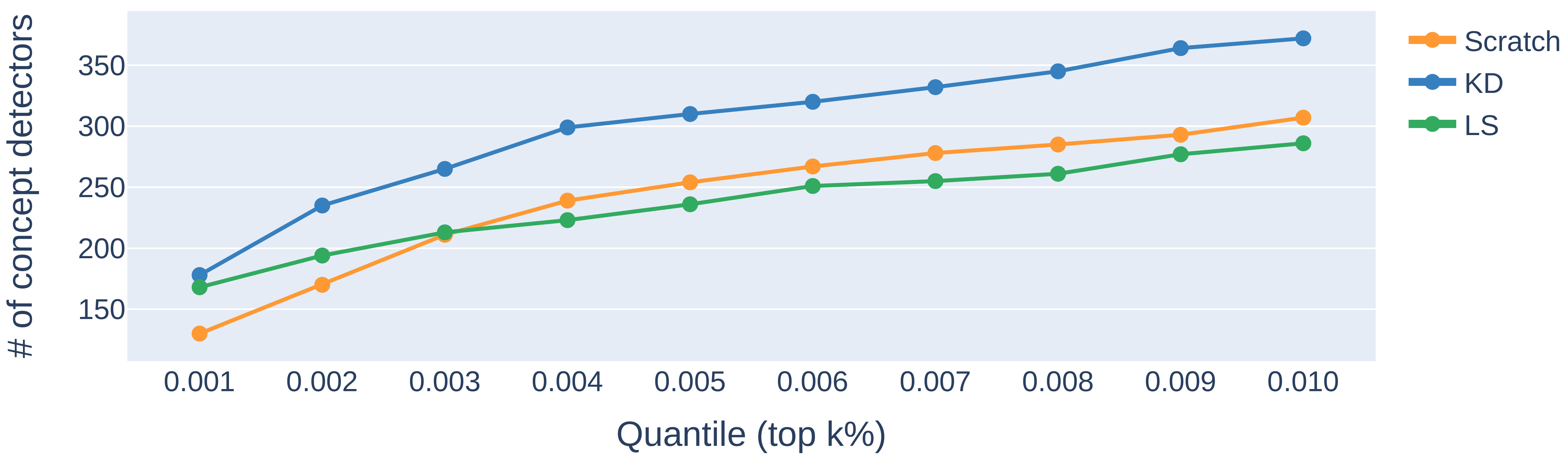}
    \caption{The total number of concept detectors of $f_{\mathrm{scratch}}$, $f_{\mathrm{KD}}$ and $f_{\mathrm{LS}}$ for varying the quantiles (from 0.001 to 0.010 in increments of 0.001). The quantile indicates the top k\% of activation value for obtaining a $T_i$.}
    \label{fig:ablation_quantile}
    \vspace{3mm}
    \includegraphics[width=0.7\linewidth]{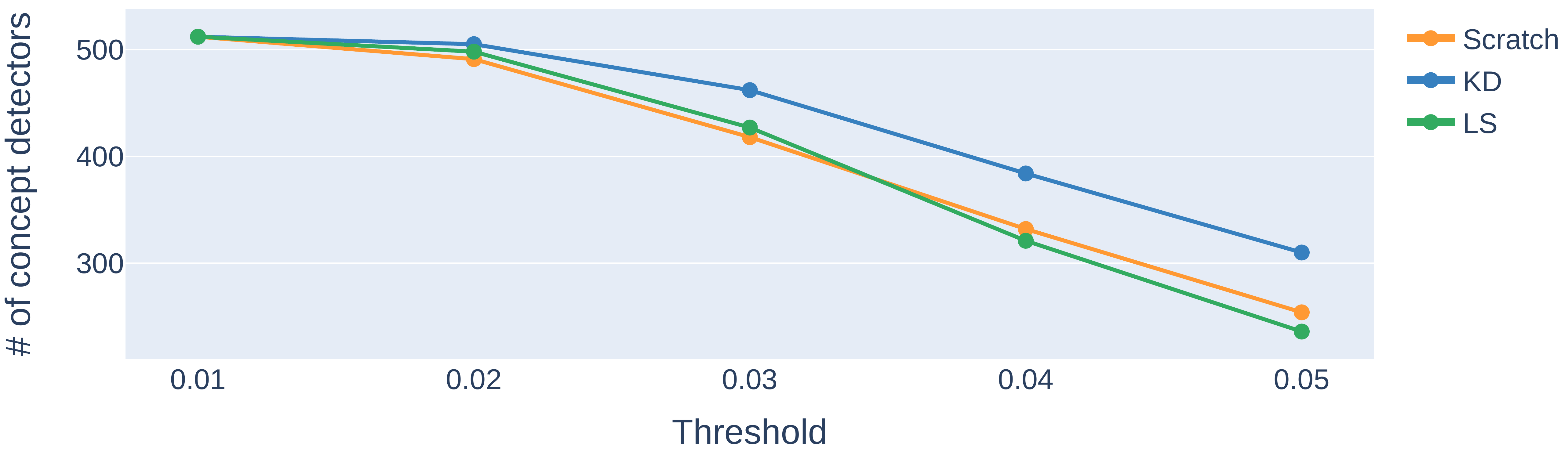}
    \caption{The total number of concept detectors of $f_{\mathrm{scratch}}$, $f_{\mathrm{KD}}$ and $f_{\mathrm{LS}}$ for varying the IoU thresholds (from 0.01 to 0.05 in increments of 0.01). If IoU score between the masked activation map and annotation mask is higher than IoU threshold, we call that unit a concept detector.}
    \label{fig:ablation_threshold}
\end{figure}

In the main paper, we conducted a comparison of model interpretability using quantile and IoU thresholds of 0.005 and 0.05.
The quantile represents the top k\% of the activation value used to obtain $T_i$ (as described in Section~\ref{sec:3.1} of the main paper).
A unit is considered a concept detector if the IoU score between the masked activation map and annotation mask exceeds the IoU threshold.
Both the quantile and IoU threshold are important hyper-parameters for obtaining the concept detectors.
Figure \ref{fig:ablation_quantile} and \ref{fig:ablation_threshold} present the results of the interpretability measurements for various quantiles and IoU thresholds.
We show that KD enhances model interpretability regardless of these two hyper-parameters.

\subsection{Model interpretability for various layers}
\label{sec:ap_A.2}
Table \ref{tab:various layers} lists the number of unique detectors and the total number of concept detectors in lower layers of $f_{\mathrm{scratch}}$, $f_{\mathrm{KD}}$ and $f_{\mathrm{LS}}$.
Each model was the same model presented in Table 1 of our main paper.
The quantile and IoU threshold were also equal to 0.005 and 0.05.
We demonstrate that $f_{\mathrm{KD}}$ had more interpretable units, even in the lower layers.
In particular, the interpretability gap with $f_{\mathrm{scratch}}$ widened from layer 3.

\begin{table}[t]
\centering
\caption{The comparison of model interpretability of $f_{\mathrm{scratch}}$, $f_{\mathrm{KD}}$ and $f_{\mathrm{LS}}$ in the lower layers; We obtained consistent results with the comparison of model interpretability of the main paper.}
\resizebox{0.55\linewidth}{!}{%
\begin{tabular}{c|ccc|ccc}\toprule
        & \multicolumn{3}{c|}{Unique}             & \multicolumn{3}{c}{Total}              \\
Model   & Layer1     & Layer2      & Layer3      & Layer1     & Layer2      & Layer3      \\\midrule
Scratch & 0          & 7           & 30          & 0          & 7           & 37          \\
LS      & 1          & 8           & 29          & 1          & 9           & 35          \\
KD      & \textbf{1} & \textbf{10} & \textbf{39} & \textbf{1} & \textbf{10} & \textbf{52}\\\bottomrule
\end{tabular}%
}
\label{tab:various layers}
\end{table}

\subsection{Model interpretability for various architectures}
\label{sec:ap_A.3}
In the main paper, we focused a single setup for comparing the model interpretability with ResNet-18.
It is essential to show that KD enhances model interpretability for architectures other than ResNet-18. 
We additionally measured interpretability of four different architectures.
Table \ref{tab: various architectures} present the interpretability for various architectures.
Regardless of the model architecture, KD enhances the interpretability of models.
In the case of MobileNet\_v2, model interpretability was further improved when the teacher was MobileNet\_v2 than the teacher was ResNet-34.
We obtain the consistent results with the measurement of the main paper that self-KD enhances the model interpretability better than vanilla KD.
\begin{table}[tp]
\centering
\caption{The comparison of model interpretability with various architectures; R, M, and E in the first and second rows represent ResNet~\citeAppendix{ap_he2016deep}, MobileNet~\citeAppendix{ap_howard2017mobilenets}, and EfficientNet~\citeAppendix{ap_tan2019efficientnet}, respectively. The symbol ``-'' indicates that the model was trained from scratch, not KD.}
\vspace{1mm}
\resizebox{0.8\linewidth}{!}{%
\begin{tabular}{cccccccccc}\toprule
Teacher & - & R-50 & - & R-50 & -  & R-34 & M\_v2 & -  & E\_b2 \\
Student & R-34 & R-34 & R-50 & R-50 & M\_v2 & M\_v2 & M\_v2 & E\_b2 & E\_b2 \\\midrule
Unique&161&\textbf{185}&602&\textbf{685}&244&351&\textbf{393}&96&\textbf{469}\\
Total&267&\textbf{329}&1008&\textbf{1196}&348&655&\textbf{694}&129&\textbf{882} \\\bottomrule              
\end{tabular}
}
\label{tab: various architectures}
\end{table} 

\section{Experimental details}
\subsection{Experimental environment}
\label{sec:ap_B.1}
We conduct all experiments introduced in the main paper with the following environments.
\begin{enumerate}

    \item \textbf{CPU}: Intel(R) Xeon(R) Gold 6258R
    \item \textbf{GPU}: NVIDIA A40 48GB GDDR6
    \item \textbf{CUDA version}: 11.4
    \item \textbf{Library}: PyTorch~\citeAppendix{ap_paszke2019pytorch}
\end{enumerate}

\subsection{Experimental setup for training $f_{\mathrm{scratch}}$, $f_{\mathrm{KD}}$, and $f_{\mathrm{LS}}$}
\label{sec:ap_B.2}
The detailed experimental settings for training $f_{\mathrm{scratch}}$, $f_{\mathrm{KD}}$, and $f_{\mathrm{LS}}$ are as follows:
All models were trained on the ImageNet dataset~\citeAppendix{ap_russakovsky2015imagenet}, and the total number of epochs was 100.
We performed various KD experiments using the TorchDistill library~\citeAppendix{ap_matsubara2021torchdistill}.
For the teacher model, we used the pre-trained ResNet-34 architecture provided by Torchvision~\citeAppendix{ap_paszke2017automatic}.
We used the ResNet-18 architecture for $f_{\mathrm{KD}}$ and the same architecture for $f_{\mathrm{scratch}}$ and $f_{\mathrm{LS}}$ for an unbiased comparison.
We used SGD optimization as the optimizer.
We set the initial learning rate to 0.1 and divided it by ten every 30, 60, and 90 epoch for scheduling.
We set the temperature to four. 
$\alpha$, a hyper-parameter to determine the ratio of the correct answer to $\boldsymbol{z}_t$, was set to 0.5 for training.
For the LS, we set the $\alpha$ value to 0.1.
We report the results of the hyper-parameter setting with the highest generalization performance (top-1 test accuracy).

\subsection{Experimental setup for the entropy measurement experiments}
\label{sec:ap_B.3}
To show that class-similarity information is actually transferred from teacher to student model via logit distillation, we compared the entropy measured from entire and within same category of $f_{\mathrm{scratch}}$, $f_{\mathrm{KD}}$, and $f_{\mathrm{LS}}$ in Section~\ref{sec:4.1}. 
For the entropy measurement experiment, we used the pre-trained ResNet-34 architecture provided by Torchvision as a teacher model.
We used the ResNet-18 architecture for $f_{\mathrm{scratch}}$, $f_{\mathrm{KD}}$, and $f_{\mathrm{LS}}$.
We used $f_{\mathrm{scratch}}$ as a pre-trained model provided by PyTorch.
The hyper-parameters of $f_{\mathrm{KD}}$ were $\alpha = 0.5$, $T = 1$, and $\alpha$ of the $f_{\mathrm{LS}}$ was 0.5.
The other setting (e.g., the total number of epochs, optimization, and learning rate) were the same as in the previous section.
\subsection{Experimental setup for various KD methods}
\label{sec:ap_B.4}
In the main paper, we compared the interpretability of the models trained with various KD methods (in Table 1).
We describe the experimental setup of models trained with various KD.
We conducted most of KD experiments using the TorchDistill library. 
For all experiments except self-KD, we used ResNet-34 architecture as teachers and ResNet-18 architecture as students.
\subsubsection{Attention transfer (AT) \citeAppendix{ap_zagoruyko2016paying}}
\begin{itemize}
    \item[$\bullet$] \textbf{training epochs}: 100
    \item[$\bullet$] \textbf{batch size}: 256
    \item[$\bullet$] \textbf{attention pair}: layer3, and layer4
    \item[$\bullet$] \textbf{attention loss factor}: 1,000
\end{itemize}

\subsubsection{Factor transfer (FT)~\citeAppendix{ap_kim2018paraphrasing}}
\begin{itemize}
    \item[$\bullet$] \textbf{training epochs for paraphraser}: 1
    \item[$\bullet$] \textbf{number of input channels for paraphraser and translator}: 512
    \item[$\bullet$] \textbf{number of output channels for paraphraser and translator}: 256
    \item[$\bullet$] \textbf{training epochs for student}: 90
    \item[$\bullet$] \textbf{batch size}: 256
    \item[$\bullet$] \textbf{norm type}: 1
    \item[$\bullet$] \textbf{transferred layer}: layer4
    \item[$\bullet$] \textbf{factor transfer loss factor}: 1,000
\end{itemize}

\subsubsection{Contrastive representation distillation (CRD)~\citeAppendix{ap_tian2019contrastive}}
\begin{itemize}
    \item[$\bullet$] \textbf{training epochs}: 100
    \item[$\bullet$] \textbf{batch size}: 85
    \item[$\bullet$] \textbf{number of negative samples}: 16384
    \item[$\bullet$] \textbf{feature dimension}: 128
    \item[$\bullet$] \textbf{temperature for contrastive learning}: 0.07
    \item[$\bullet$] \textbf{momentum}: 0.5
    \item[$\bullet$] \textbf{contrastive loss factor}: 0.8
\end{itemize}

\subsubsection{Self-supervision knowledge distillation (SSKD)~\citeAppendix{ap_xu2020knowledge}}
\begin{itemize}
    \item \textbf{training teacher SS module}
    \begin{itemize}
        \item[$\bullet$] \textbf{training epochs}: 30
        \item[$\bullet$] \textbf{batch size}: 85
        \item[$\bullet$] \textbf{feature dimension}: 512
        \item[$\bullet$] \textbf{optimizer}: SGD
        \item[$\bullet$] \textbf{learning rate}: 0.1 divided by 10 at 10, 20 epoch
    \end{itemize}
    \item \textbf{training student}
    \begin{itemize}
        \item[$\bullet$] \textbf{training epochs}: 100
        \item[$\bullet$] \textbf{feature dimension}: 512
        \item[$\bullet$] \textbf{KD temperature}: 4.0
        \item[$\bullet$] \textbf{SS temperature}: 0.5
        \item[$\bullet$] \textbf{TF temperature}: 4.0
        \item[$\bullet$] \textbf{SS ratio}: 0.75
        \item[$\bullet$] \textbf{TF ratio}: 1.0
        \item[$\bullet$] \textbf{loss weights [CE, KD, SS, TF]}: [1.0, 0.9, 10.0, 2.7]
    \end{itemize}
\end{itemize}

\subsubsection{Self knowledge distillation (Self-KD)~\citeAppendix{ap_furlanello2018born}}
\begin{itemize}
    \item[$\bullet$] \textbf{teacher architecture}: ResNet-18
    \item[$\bullet$] \textbf{training epochs}: 100
    \item[$\bullet$] \textbf{batch size}: 256
    \item[$\bullet$] \textbf{learning rate}: 0.1 divided by 0.1 at 30, 60, 90 epoch
    \item[$\bullet$] \textbf{momentum}: 0.9
    \item[$\bullet$] \textbf{$\boldsymbol{\alpha}$}: 0.1
    \item[$\bullet$] \textbf{$\boldsymbol{T}$}: 4.0
\end{itemize}
\subsection{Experimental setup for KD using $f_{\mathrm{LS}}^{\mathrm{teacher}}$}
\label{sec:ap_B.5}
In addition to the effect of the presence or absence of class-similarity information on model interpretability, we also analyzed how the degree of transferred class-similarity information affects the model interpretability in Section~\ref{sec:4.3} of the main paper.
This section presents the detailed experimental settings for KD using $f_{\mathrm{LS}}^{\mathrm{teacher}}$ shown in Figure~\ref{fig:LS_trained}.
We used the ResNet-34 architecture for $f_{\mathrm{LS}}^{\mathrm{teacher}}$, and $f_{\mathrm{LS}}^{\mathrm{teacher}}$ is trained using LS with $\alpha = 0.1$.
For the student model, we used the ResNet-18 architecture and we set the $\alpha = 0.5$ for KD.
We varied the temperature to 1, 2, and 4 to analyze the impact of transferred class-similarity information on the model interpretability.
We trained each model thrice based on different initial points to avoid variations.
The other settings (e.g., the total number of epochs, optimization, and learning rate) were the same as in Section~\ref{sec:ap_B.2}.   

\subsection{Experimental setup for synthesized dataset experiments}
\label{sec:ap_B.6}
In the main paper, we demonstrate KD enhances the model interpretability with various notions and dataset (in Section~\ref{sec:5}).
We describe the setup of experiments on the synthesized dataset. 
For experiments on synthesized dataset, we used the ResNet-34 architecture without pre-training provided by Torchvision for a teacher model. 
We used the ResNet-18 architecture for a student model.
We used the same architecture for the models trained from scratch.
Since the number of classes for the synthesized dataset is 10, we added one fully connected (FC) layer to the ResNet backbone, and the output of FC layer is 10.
Because the dataset is not complicated and overfitting easily occurs, as suggested by~\citeAppendix{ap_tjoa2020quantifying}, we trained the models with small epochs (training epoch = 4), and the number of batch size was 4.
We set the first learning rate to 0.001 with 0.00005 weight decay.
Adam optimization was used as the optimizer.
For KD training, $\alpha$ and $T$ were set to 0.5 and 4, respectively.
We used the Saliency function of the Captum library to get the saliency map of the models for evaluations~\citeAppendix{ap_kokhlikyan2020captum}.

\subsection{Experimental setup for calculating DiffROAR}
\label{sec:ap_B.7}
In the main paper, we measure the DiffROAR scores (in Section~\ref{sec:5.2}).
DiffROAR is the difference in the predictive power of datasets, with the top-k\% and bottom-k\% of pixels removed by ordering the feature attribution of the model.
We present the experimental setup for calculating DiffROAR scores.
We used the Saliency function of the Captum library to obtain the feature attribution of the model.
We used ResNet-18 as the teacher and student (Self-KD).
For the CIFAR dataset, $\alpha$ and $T$ were set to 0.1 and one.
We re-trained the ResNet-18 model for top-k\% and bottom-k\% removed datasets with 60 epochs.
For MNIST, we set $\alpha$ and $T$ to 0.1 and four.
We re-trained the ResNet-18 model for top-k\% and bottom-k\% removed datasets with 10 epochs.
The differences in accuracy for top-k\% and bottom-k\% are presented in Table 4 of the main paper.

\subsection{Experimental setup for NLP distillation}
\label{sec:ap_B.8}
In the main paper, we demonstrate that KD enhances model interpretability in NLP tasks.
We conducted an experiment using BERT for a classification task and utilized the Standard Sentiment Treebank (SST) dataset to measure model interpretability.
The original SST dataset comprises five classes (`very negative', `negative', `neutral', `positive', and `very positive'). 
However, we train the model using only four classes (`very negative', `negative', `positive', and `very positive') because unlike the negative and positive classes, the `neutral' class does not contain similarity information with other classes.
To perform distillation, we set the values of $\alpha$ and $T$ to 0.5 and four, respectively. 
We trained the BERT-student model with three and six layers for 20 epochs. 

\section{Pseudocode of obtaining concept detectors}
\label{sec:ap_C}
We present the pseudocode of obtaining the concept detector to facilitate the understanding of network dissection, and the code is shown in Algorithm \ref{alg:psudo}.
\begin{algorithm}
\caption{Obtaining the concept detectors}\label{alg:psudo}
\textbf{Require}: Broden dataset $X$, target model $f$, and target concept $c$
\begin{algorithmic}[1]
\STATE $N$ $\gets$ the number of convolutional units in fourth layer of $f$
\FOR {$x$ $\in R^{n \times n}$ in $X$}
    \FOR {$i = 1,2, ..., N$}
        \STATE Collect the activation map $A_i(x) \in R^{d \times d}$, where $d < n$
    \ENDFOR
\ENDFOR
\STATE $a_i \gets$ the distribution of individual unit activation
\FOR {$x$ $\in R^{n \times n}$ in $X$}
    \FOR {$i = 1,2, ..., N$}
        \STATE Calculate $T_i$ to satisfy $P(a_i \geq T_i) = 0.005$ 
        \STATE Interpolate $A_i(x)$ to be $\in R^{n \times n}$
        \STATE $A_i(x)$ $\gets$ $A_i(x) \geq T_i$
        \STATE $M_c(x)$ $\gets$ annotation mask of $x$ for concept $c$
        \STATE Compute $IoU_{i,c}$ value between $A_i(x)$ and $M_c(x)$
        \IF{$IoU_{i,c}$ $\geq$ $0.05$:}
            \STATE Unit $i$ is the concept detector of the concept $c$
        \ENDIF
    \ENDFOR
\ENDFOR
\end{algorithmic}
\end{algorithm}

\begin{table}[bp]
\centering
\caption{Comparison of entropy within two similar classes (\texttt{komondor} and \texttt{old English sheepdog})}
\vspace{1mm}
\resizebox{0.4\linewidth}{!}{%
\begin{tabular}{cc}\toprule
Model                  & Entropy \\\midrule
Scratch                & 0.944   \\
Knowledge distillation & \textbf{0.953}   \\
Label smoothing        & 0.872   \\\bottomrule
\end{tabular}
}
\label{tab:entropy}
\end{table}

\section{The example samples of synthesized dataset}
\label{sec:ap_D}
\begin{figure}[h]
    \centering
    \includegraphics[width=0.6\linewidth]{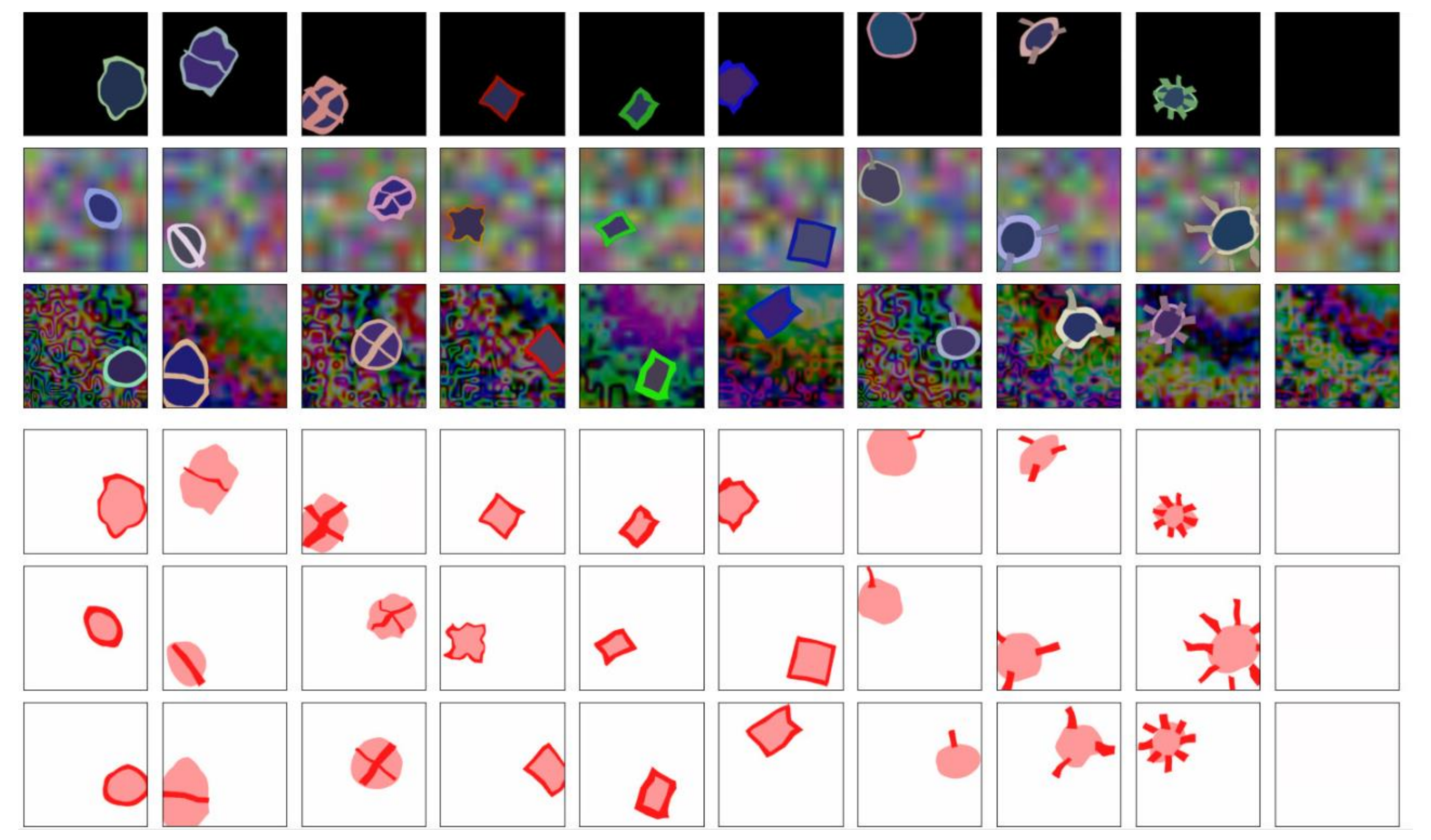}
    \caption{Samples and ground truths of the synthetic dataset that we generated. Row 1-3: Sample images of the 10 classes. Row 4-6: Ground truths of the sample images. The synthesized dataset formed a hierarchy with a circular (columns 1 to 3), rectangular (columns 4 to 6), and tail category (columns 7 to 9). Samples of the last column is a class with no objects. Each sample exhibited one among the three types (dark, blurred, and noisy) of random background.}
    \label{fig:synthetic dataset sample}
\end{figure}

To verify that KD improves model interpretability except for the Broden dataset and the number of concept detectors, we present the result of five-band-scores and radar plots using the synthesized dataset in Section~\ref{sec:5.1} of the main paper.
This section presents the example samples of synthesized that we generated.
The synthesized dataset has the ground truth for the heatmap and was proposed by~\citetAppendix{ap_tjoa2020quantifying}.  
The synthesized dataset comprised 10 classes, and examples for each class and the ground truth that we generated are shown in Figure~\ref{fig:synthetic dataset sample}.
We generated 6,400 training and 1,600 test samples.
The first three rows represent the ground truths of the sample images. 
The synthesized dataset formed a hierarchy with a circular (columns 1 to 3), rectangular (columns 4 to 6), and tail category (columns 7 to 9).
Each sample exhibited one among the three types (dark, blurred, and noise) of random background.
The ground truth of the synthesized dataset has three regions: class 0), a background that does not contain any classification information, shown as a white region; class 1), localization information, the location of an object, shown as a light pink region; and class 2), the distinguishing feature, which is crucial for distinguishing between classes, is shown as a dark pink area.

\section{Additional experimental results}
\subsection{Qualitative results of entropy measurement experiments}
\label{sec:ap_E.1}
\begin{figure}[h]
    \centering
    \includegraphics[width=0.55\linewidth]{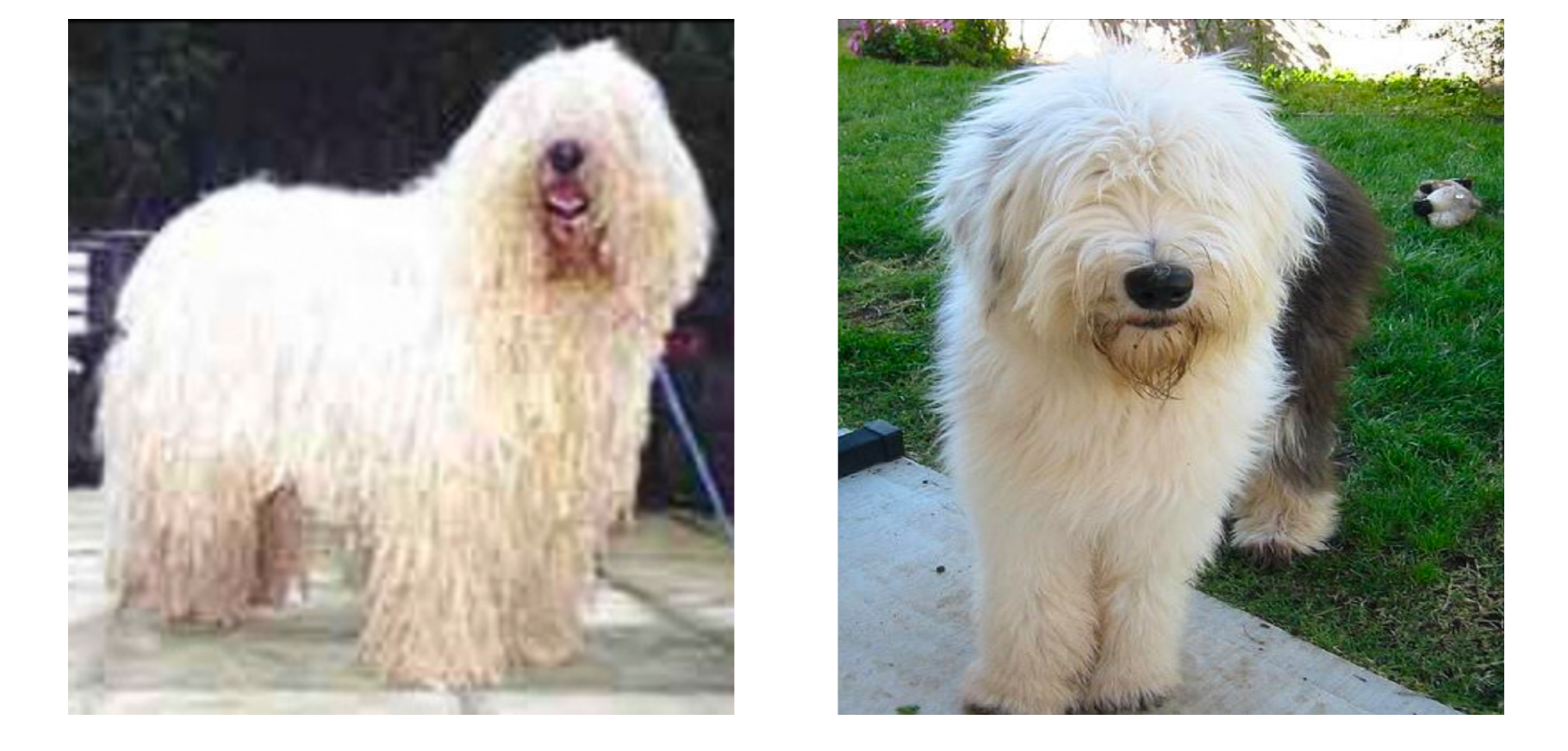}
    \caption{Example images of \texttt{komondor} (left) and \texttt{old English sheepdog} (right).}
    \label{fig:qualitative}
\end{figure}
In this section, we demonstrate the entropy measured in two classes with high similarity to show that KD contains class-similarity information well.
Two classes with high similarity are \texttt{komondor} and \texttt{old English sheepdog} belonging to ``sheepdog'' category, and we present the example image in Figure \ref{fig:qualitative}.
Both the \texttt{komondor} and the \texttt{old English sheepdog} are dogs with their faces covered in hair, with the difference that the former has a white fur, whereas the latter has a grayish fur on the back of the body.
We obtained the output distribution of $f_{\mathrm{scratch}}$, $f_{\mathrm{KD}}$ and $f_{\mathrm{LS}}$ when the correct answer class was \texttt{komondor} or \texttt{old English sheepdog} as the input sample.
Table \ref{tab:entropy} lists the results of measuring entropy values using the output logit values of the two classes when all models had the correct answer.
Even for two classes with high similarity, the entropy of $f_{\mathrm{KD}}$ was the largest, and the entropy of $f_{\mathrm{LS}}$ decreased significantly compared to $f_{\mathrm{scratch}}$.
Through qualitative entropy measurement experiments, we confirmed that the $f_{\mathrm{KD}}$ contained class-similarity information well, but not the $f_{\mathrm{LS}}$.

Table \ref{tab:category} presents the number of classes for each category used in the entropy measurement experiment in the main paper.
We divided 1,000 classes into 67 categories based on the coarse ImageNet category classification proposed by~\citeAppendix{ap_eshed_novelty_detection}.
Among the 67 categories, we excluded the classes of ``other'' category because we could not state that similar classes were grouped together in that category.

\subsection{Visualizations of loss gradient for MNIST dataset}
\label{sec:ap_E.2}
\begin{figure}[bp]
    \centering
    \includegraphics[width=0.64\linewidth]{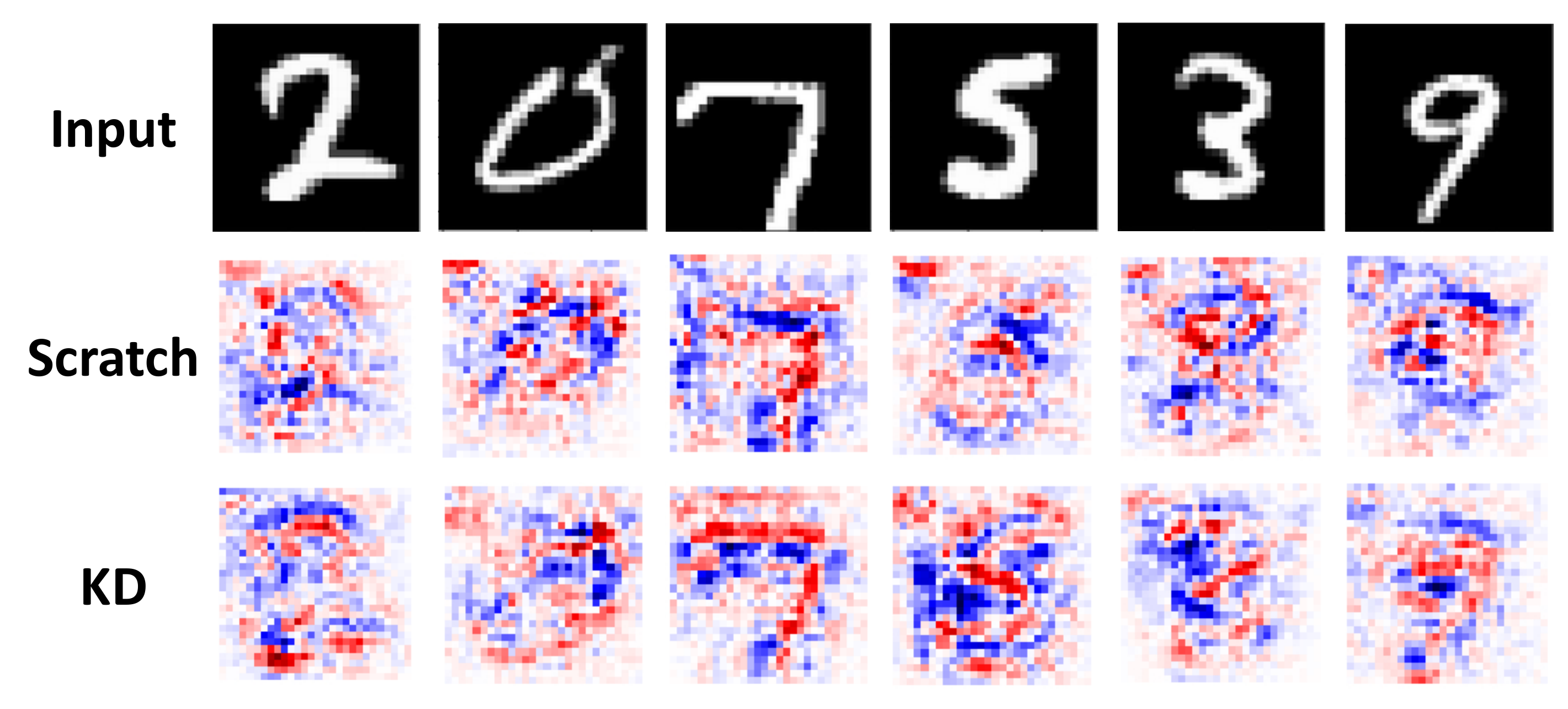}
    \caption{Visualization of loss gradients of $f_{\mathrm{scratch}}$ and $f_{\mathrm{KD}}$ on the testset of MNIST.}
    \label{fig:MNIST loss gradient}
\end{figure}

We presented visualization of the loss gradients on the ImageNet in the Section~\ref{sec:5.3} of the main paper.
In addition to ImageNet, we present the visualization of the loss gradients of $f_{\mathrm{scratch}}$ and $f_{\mathrm{KD}}$ on the MNIST dataset, and the results are in Figure \ref{fig:MNIST loss gradient}.
The gradients of $f_{\mathrm{KD}}$ were more aligned with the semantic important regions (region of the numbers) than $f_{\mathrm{scratch}}$.
We demonstrate that $f_{\mathrm{KD}}$ learned more human-perceptually relevant features than $f_{\mathrm{scratch}}$ for various datasets.
\begin{table}[bp]
\vspace{-1mm}
\caption{Comparison of model interpretability of $f_{\mathrm{scratch}}$ and $f_{\mathrm{KD}}$ on the SST dataset with various layers of student model (the higher the better); the subscript refers to the layers of the student model}
\centering
\vspace{1mm}
\resizebox{0.5\linewidth}{!}{%
    \begin{tabular}{c| c  c  c }  
    \toprule
        Model                   &  Accuracy        & AUROC          & AUPRC  \\\midrule
        $f_{\textrm{scratch}}^{\textrm{3-layer}}$  &  0.677           & 0.689          & 0.810     \\
        $f_{\textrm{KD}}^{\textrm{3-layer}}$       &  \textbf{0.722}  & \textbf{0.720} & \textbf{0.831} \\\midrule
        $f_{\textrm{scratch}}^{\textrm{6-layer}}$  &  0.668           & 0.670          & 0.723     \\
        $f_{\textrm{KD}}^{\textrm{6-layer}}$       &  \textbf{0.722}  & \textbf{0.793} & \textbf{0.829} \\
        \bottomrule
    \end{tabular}
    }
    \label{tab:Appendix_NLP distillation}
\end{table}
\begin{table}[tp]
\centering
\caption{The number of classes belonging to each category}
\resizebox{0.95\linewidth}{!}{%
\begin{tabular}{cc | cc | cc}\toprule
Category       & \# of classes & Category   & \# of classes & Category             & \# of classes    \\\midrule
arachnid       & 8                 & mollusk    & 6                 & building             & 37                   \\
armadillo      & 1                 & mongoose   & 3                 & clothing             & 47                   \\
bear           & 5                 & monotreme  & 2                 & container            & 18                   \\
bird           & 59                & person     & 4                 & cooking              & 27                   \\
bug            & 25                & plant      & 3                 & decor                & 22                   \\
butterfly      & 6                 & primate    & 19                & electronics          & 49                   \\
cat            & 4                 & rabbit     & 3                 & fence                & 3                    \\
coral          & 5                 & rodent     & 7                 & food                 & 28                   \\
crocodile      & 2                 & salamander & 5                 & furniture            & 34                   \\
crustacean     & 9                 & shark      & 4                 & hat                  & 8                    \\
dinosaur       & 1                 & sloth      & 2                 & instrument           & 28                   \\
dog            & 119               & snake      & 16                & lab equipment        & 2                    \\
echinoderms    & 3                 & trilobite  & 1                 & other                & 19                   \\
ferret         & 7                 & turtle     & 5                 & outdoor scene        & 32                   \\
fish           & 13                & ungulate   & 16                & paper                & 9                    \\
flower         & 5                 & vegetable  & 7                 & sports equipment     & 13                   \\
frog           & 3                 & wild cat   & 9                 & technology           & 27                   \\
fruit          & 14                & wild dog   & 11                & tool                 & 43                   \\
fungus         & 7                 & accessory  & 18                & toy                  & 4                    \\
hog            & 3                 & aircraft   & 5                 & train                & 4                    \\
lizard         & 11                & ball       & 9                 & vehicle              & 49                   \\
marine mammals & 4                 & boat       & 15                & weapon               & 10                   \\
marsupial      & 3                 &            &                   & \multicolumn{1}{l}{} & \multicolumn{1}{l}{} \\\bottomrule
\end{tabular}%
}
\label{tab:category}
\end{table}

\subsection{BERT model interpretability for various layers}
\label{sec:ap_E.3}
We demonstrated the model interpretability using the BERT model in Section~\ref{sec:5.4}.
The SST dataset provides a label for each word as either positive or negative, serving as the ground truth for saliency (attribution), similar to the synthesized dataset in the main paper.
We computed Integrated Gradients (IG) attribution scores from the validation and test samples using the LayerIntegratedAttribution function of the Captum library~\citeAppendix{kokhlikyan2020captum}.
Accuracy, AUROC, and AUPRC were listed as measures of model interpretability in Table~\ref{tab:NLP distillation}.
In Table~\ref{tab:NLP distillation}, we presented only the experimental results of applying KD with a 12-layer BERT as the teacher model and a 3-layer BERT as the student model. 
In Table~\ref{tab:Appendix_NLP distillation}, we show the model interpretability when varying the layers of the student model.
Our results show that KD enhances model interpretability, even when the layers of the student model are varied.

\newpage
\bibliographystyleAppendix{icml2022}
\bibliographyAppendix{supp}
\end{document}